\documentclass[sigconf,noend]{acmart}
\usepackage{algorithm}
\usepackage{algpseudocode}
\usepackage{multirow}
\usepackage{graphicx} 
\usepackage{placeins}

\usepackage{tikz}
\usepackage{rotating}  
\usepackage{caption}
\usepackage[many]{tcolorbox}
\usepackage{booktabs}
\definecolor{sysPink}{RGB}{255, 229, 217}   
\definecolor{varGreen}{RGB}{226, 240, 217}  
\definecolor{respBlue}{RGB}{218, 232, 252}  
\definecolor{headGray}{RGB}{80, 80, 80}     
\definecolor{bodyGray}{RGB}{250, 250, 250} 
\definecolor{exYellow}{RGB}{255, 250, 205}
\newif\ifanonymous

\newtcbox{\highlightbox}[1][white]{
    on line, 
    arc=3pt, outer arc=3pt,
    colback=#1, 
    colframe=#1!80!black,
    boxsep=0pt, left=3pt, right=3pt, top=2pt, bottom=2pt,
    boxrule=0.5pt,
    fontupper=\small\ttfamily 
}

\newtcolorbox{promptbox}[2][]{
    enhanced,
    title={#2},
    colback=bodyGray,
    colframe=headGray,
    coltitle=white,
    fonttitle=\bfseries\normalsize, 
    arc=5pt, 
    boxrule=1pt,
    left=8pt, right=8pt, top=8pt, bottom=8pt,
    fontupper=\small, 
    #1
}

\AtBeginDocument{%
  }

\usepackage{algpseudocode}

\AtBeginDocument{%
  }

\copyrightyear{2026}
\acmYear{2026}
\setcopyright{cc}
\setcctype{by}
\acmConference[MM '26]{Proceedings of the 34th ACM International Conference on Multimedia}{November 10--14, 2026}{Rio de Janeiro, Brazil}
\acmBooktitle{Proceedings of the 34th ACM International Conference on Multimedia (MM '26), November 10--14, 2026, Rio de Janeiro, Brazil}
\acmDOI{10.1145/3767308.3836133}
\acmISBN{979-8-4007-2213-4/2026/11}

\begin{document}

\title{ViSR-KGC: Visual Subgraph Reasoning with Vision-Language Models for Multimodal Knowledge Graph Completion}

\author{Jiafan Li}
\authornote{Both authors contributed equally to this research.}
\affiliation{%
  \institution{Institute of Software, Chinese Academy of Sciences, Beijing, China}
  \institution{University of Chinese Academy of Sciences, Beijing, China}
  \institution{CITIC Securities, Beijing, China}
  \country{}
  }

\author{Mengxue Yang}
\authornotemark[1]
\affiliation{%
  \institution{University of Chinese Academy of Sciences, Beijing, China}
  \institution{Institute of Software, Chinese Academy of Sciences, Beijing, China}
  \country{}
  }

\author{Jiaqi Zhu}
\correspondingauthor
\affiliation{%
  \institution{Institute of Software, Chinese Academy of Sciences, Beijing, China}
  \institution{University of Chinese Academy of Sciences, Beijing, China}
  \country{}
  }
\email{zhujq@ios.ac.cn}

\author{Liang Chang}
\affiliation{%
  \institution{School of Artificial Intelligence, Beijing Normal University, Beijing, China}
  \country{}
  }

\author{Ying Li}
\affiliation{%
  \institution{University of Chinese Academy of Sciences, Beijing, China}
  \country{}
  }

\author{Hongan Wang}
\affiliation{%
  \institution{Institute of Software, Chinese Academy of Sciences, Beijing, China}
  \institution{University of Chinese Academy of Sciences, Beijing, China}
  \country{}
  }

\renewcommand{\shortauthors}{Jiafan Li et al.}

\begin{abstract}
Knowledge graph completion (KGC) aims to infer missing entities or relations from incomplete graph structures, and has evolved into multimodal knowledge graph completion (MMKGC), where entities are associated with multiple modalities such as text and images. Traditional representation learning approaches follow the embedding-based paradigm and may struggle when relation-specific evidence is limited. Meanwhile, LLM-based reasoning methods typically linearize graph structures into textual prompts, which obscures structural topology and neglects vital visual information.
While vision-language models (VLMs) excel at multimodal reasoning, they cannot natively interpret structured graph topology, particularly when it comes to knowledge graphs where nodes and edges carry complex semantics.
To bridge this gap, we propose ViSR-KGC, a visual subgraph reasoning approach for KGC. It integrates three complementary capabilities to capture semantic correlations: identifying global topology dependencies via representation learning, analyzing local multimodal evidence using VLMs, and providing necessary commonsense knowledge inherent in pre-trained models.
Based on learned multimodal embeddings, our framework first extracts a compact and query-aware subgraph from the MMKG.
Then, this subgraph is transformed into a visually interpretable image using a layout strategy selected through empirical comparison. Finally, the visualized subgraph, entity images, textual descriptions, and candidate answers are combined into a unified prompt, enabling the VLM to infer the missing entity.
Experimental results on two real-world MMKG datasets demonstrate that ViSR-KGC consistently outperforms traditional embedding-based models and state-of-the-art multimodal baselines. This highlights the unique advantages of jointly leveraging VLMs' visual perception and internal knowledge for structured reasoning, thereby paving the way toward deeper investigation into VLMs' capacity for semantic graph analysis.
The full version including Supplementary Materials is available at \url{https://arxiv.org/abs/2608.05833}.
\end{abstract}





\maketitle

\begin{figure}[t]
    \centering
    \includegraphics[width=0.8\linewidth]{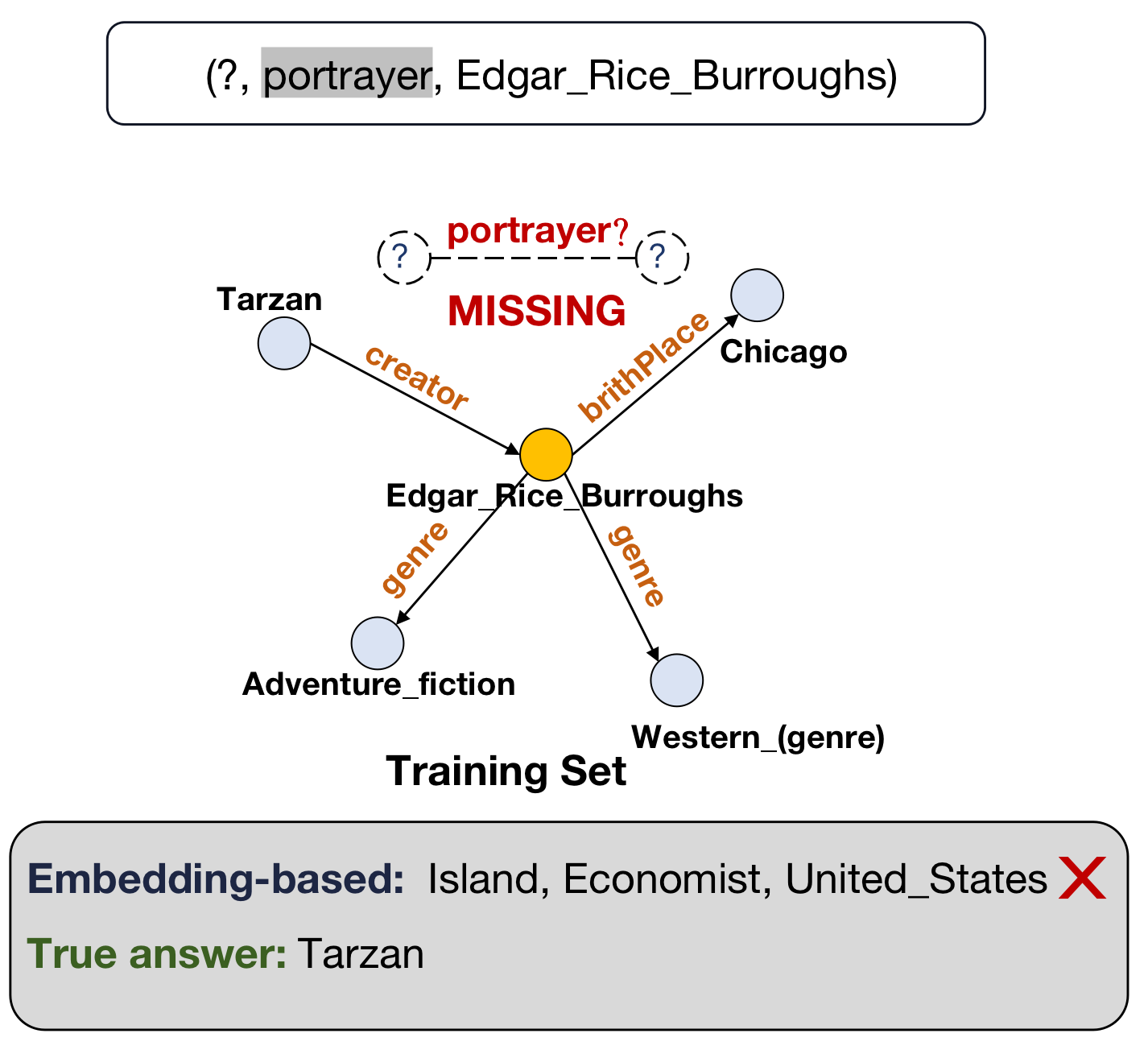}
    \caption{Incorrect embedding-based prediction for a relation type with limited evidence.}
    \label{fig:intro}
\end{figure}

\section{Introduction}

Knowledge graphs (KGs) represent real-world knowledge in a structured form by organizing entities and relations into factual triples, 
widely used in
semantic search~\cite{dong2014knowledgevault}, recommendation systems~\cite{DBLP:conf/kdd/Wang00LC19}, and question answering~\cite{DBLP:conf/acl/SaxenaTT20}. With the rapid evolution of the Internet, some KGs now incorporate image attributes
in addition to textual content. This enrichment of node information gives rise to Multimodal Knowledge Graphs (MMKGs), which support more accurate reasoning and intuitive visualizations.


Nevertheless, real-world KGs are often incomplete, as new entities and relations continuously emerge. \emph{Knowledge graph completion} (KGC), also known as link prediction, thus becomes a fundamental task: given an incomplete triple such as $(h, r, ?)$ or $(?, r, t)$, the goal is to infer the missing entity that best satisfies the relation.
For \emph{multimodal knowledge graph completion} (MMKGC), most existing methods improve prediction accuracy by leveraging multimodal semantic information and graph structure to learn entity and relation representations \cite{10.5555/3172077.3172327,10.1145/3474085.3475470,wang2024mvtucker,zhang2024native}. However, they still follow the traditional embedding-based paradigm and thus struggle with few-shot or even zero-shot relation types due to the lack of sufficient evidence. 
As illustrated in Figure~\ref{fig:intro}, for the \textit{portrayer} relation that does not appear in the training set, embedding-based approaches cannot properly score such triples, leading to incorrect predictions.

Recently, large language models (LLMs) have been investigated for knowledge graph reasoning, harnessing their strong understanding and generative capabilities to infer missing entities via structural prompt injection. Although demonstrating promising performance, they primarily operate on textual representations of knowledge graphs and fail to incorporate entities' image attributes. Moreover, the reasoning process over graph structures is conducted in an implicit manner, making it hard to capture complex relational dependencies, especially for those associated with visual semantics. 
Meanwhile, recent advances in vision-language models (VLMs) have exhibited impressive proficiency in multimodal reasoning and visual understanding \cite{sun2026mariomultimodalgraphreasoning,DBLP:journals/corr/abs-2506-02568,lee-etal-2024-multimodal,huang2026elmmefficientlightweightmultimodal,ai-etal-2024-advancement}, but their direct application to knowledge graphs with rich semantics remains challenging and underexplored, as graph topology is essentially encoded through connectivity patterns rather than general pixel-based features. 




This raises a natural and critical question: in the VLM era, can knowledge graph completion, particularly for multimodal knowledge graphs, directly benefit from visual representations of graph structures?
Addressing this question involves three key challenges:
\textbf{(1) Subgraph extraction.} VLMs are inherently limited to processing a single image, necessitating the extraction of a compact subgraph that is highly relevant to the target query.
\textbf{(2) Subgraph visualization.} The extracted subgraph should be rendered in a suitable layout that enables VLMs to effectively identify triples, understand their correlations, and capture associated textual/visual semantics.
\textbf{(3) Prompt integration.} The prompt design must jointly incorporate local structural information from the subgraph and global features from representation learning, while harmonizing the interplay between the large structural subgraph image and small embedded entity images.


To tackle these issues, this paper proposes \textbf{ViSR-KGC}, a \textbf{visual subgraph reasoning approach for multimodal knowledge graph completion}. The key idea is to convert query-aware graph structures into visually interpretable subgraph images that can be directly processed by VLMs. By visualizing structural context, the method can not only leverage VLMs' capabilities in multimodal semantic understanding and generation, but also utilize their internal knowledge to compensate for gaps that are difficult to reason from the structured KGs alone, such as long-tail relations.
Specifically, we first obtain multimodal entity representations by integrating textual descriptions, visual features, and structural embeddings via a representation learning model, and then extract a compact subgraph consisting of entities and relations relevant to the query. After that, the subgraph is visualized using an empirically validated graph layout strategy. Finally, both the subgraph image containing entity images and textual prompts are provided to VLMs to generate the missing entity for accurate link prediction.




To sum up, this paper makes the following main contributions:
\begin{itemize}
\item To the best of our knowledge, this is the first work to \textbf{formulate MMKGC as a visual subgraph reasoning problem for VLMs}, through effectively integrating three complementary capabilities: multimodal representation learning to capture global semantics, VLM-based reasoning to extract local relational evidence, and VLMs' internal memory to provide supplementary commonsense knowledge.

\item We propose a systematic approach that extracts a query-aware subgraph based on graph structure as well as node semantics, and converts it into a visually interpretable image, facilitating VLMs to jointly reason over graph topology and multimodal entity information.

\item We conduct extensive experiments on two real-world MMKG datasets to demonstrate the effectiveness of our approach, verifying the coordinated synergy between relational reasoning and commonsense knowledge enabled by VLMs. 



\end{itemize}

\section{Related Work}

\subsection{Knowledge Graph Completion}

Knowledge graph completion (KGC) aims to predict missing
entities or relations using the structural information of
knowledge graphs. Early studies mainly adopt
\textbf{embedding-based approaches}~\cite{10.5555/2999792.2999923,10.5555/2893873.2894046,10.5555/2886521.2886624,10.1016/j.eswa.2022.119122,10.5555/3104482.3104584,DBLP:journals/corr/YangYHGD14a,10.5555/3045390.3045609,balazevic-etal-2019-tucker}, where entities and relations are
encoded as continuous vectors and optimized through scoring functions,
such as neural models ConvE~\cite{10.5555/3504035.3504256} and
ConvKB~\cite{nguyen-etal-2018-novel}.
Besides structure-only methods, a line of text-enhanced KGC
approaches incorporate entity descriptions or textualized triples
into structural reasoning, such as  StAR~\cite{DBLP:conf/www/WangSLZW021}, SimKGC~\cite{wang-etal-2022-simkgc} and CoLE~\cite{DBLP:conf/cikm/LiuSLH22}.


More recently,
\textbf{LLM-based approaches} have been explored for KGC. Methods such as
DIFT~\cite{DBLP:conf/semweb/LiuTSH24},
KICGPT~\cite{DBLP:journals/corr/abs-2402-02389},
and SLiNT~\cite{yang-etal-2025-slint}
leverage the reasoning ability of LLMs through
structural prompt injection~\cite{zhou2023large, wang-etal-2024-promise}. While these methods move
beyond pure embedding-based scoring, they still primarily operate on
textualized graph inputs, which tends to obscure graph topology and
cannot exploit multimodal entity information such as images.

\subsection{Multimodal Knowledge Graph Completion}

Multimodal knowledge graph completion (MMKGC) extends traditional KGC
by incorporating images to enrich entity representations. Most existing MMKGC methods
still follow an \textbf{embedding-based fusion paradigm}~\cite{zhang2024native,zhang2025tokenization,chen2025snag,li2025unifying}, where
structural and multimodal features are integrated into unified entity
representations like IMF~\cite{DBLP:conf/www/LiZXZX23} and HGNN-IMA~\cite{DBLP:conf/ijcai/Li00L0WYW25}.
Although these methods have achieved strong performance, the reliance on embeddings makes the reasoning process difficult to interpret. What is more, they often struggle in
few-shot and zero-shot settings, where reliable scoring patterns are
hard to learn from limited training evidence. 

Another emerging direction explores \textbf{LLM-based reasoning} for
MMKGC.
MR-MKG~\cite{lee-etal-2024-multimodal} injects multimodal knowledge into LLMs via RGAT encoding~\cite{busbridge2019relationalgraphattentionnetworks} and adapter layers.
MLaGA~\cite{DBLP:journals/corr/abs-2506-02568} aligns text, image, and graph features through a structure-aware encoder and adapts LLMs via instruction tuning.
Mario~\cite{sun2026mariomultimodalgraphreasoning} employs a graph-conditioned LLM for cross-modal alignment and a learnable router to select optimal modality configurations for LLM reasoning.
Furthermore, as a \textbf{VLM-based reasoning} method,  ELMM~\cite{huang2026elmmefficientlightweightmultimodal} devises efficient visual token compression mechanism for lightweight multimodal reasoning.
While these methods effectively leverage large models and multimodal information, they still operate on features and embeddings, leaving the graph structure implicitly encoded and thus unable to directly support visual perception of graph topology.


\section{Proposed Approach}

\subsection{Problem Formulation}
\begin{definition}
A \emph{Multimodal Knowledge Graph (MMKG)} is formally defined as a quadruple $\mathcal{G} = (\mathcal{E}, \mathcal{R}, \mathcal{T}, \mathcal{M})$, where $\mathcal{E}$ is the set of entities, $\mathcal{R}$ is the set of relation types, $\mathcal{T} \subseteq \mathcal{E} \times \mathcal{R} \times \mathcal{E}$ is the set of factual triples with each triple $(h, r, t)$ representing a directed relation $r$ from head entity $h$ to tail entity $t$, and $\mathcal{M} = \{\mathcal{M}_e\}_{e \in \mathcal{E}}$ denotes the multimodal attributes (e.g., text and image) associated with each entity $e \in \mathcal{E}$. 
\end{definition}

\begin{definition}
Given an incomplete query triple $(h,r,?)$ in MMKG $\mathcal{G}$,
the \emph{Multimodal Knowledge Graph Completion (MMKGC)} task
is to predict the most plausible tail entity $\hat{t}$
from the entity set $\mathcal{E}$:
\begin{equation}
\hat{t} =
\mathop{\text{argmax}}_{t' \in \mathcal{E}} F(h,r,t'),
\end{equation}
where $F(h,r,t')$ denotes a scoring function
that measures the plausibility of the triple $(h,r,t')$. 
Similarly, for head entity prediction $(?,r,t)$,
the predicted entity is defined as
\begin{equation}
\hat{h} =
\mathop{\text{argmax}}_{h' \in \mathcal{E}} F(h',r,t).
\end{equation}
\end{definition}

\subsection{Framework}

\begin{figure*}[t]
\centering
\includegraphics[width=\textwidth]{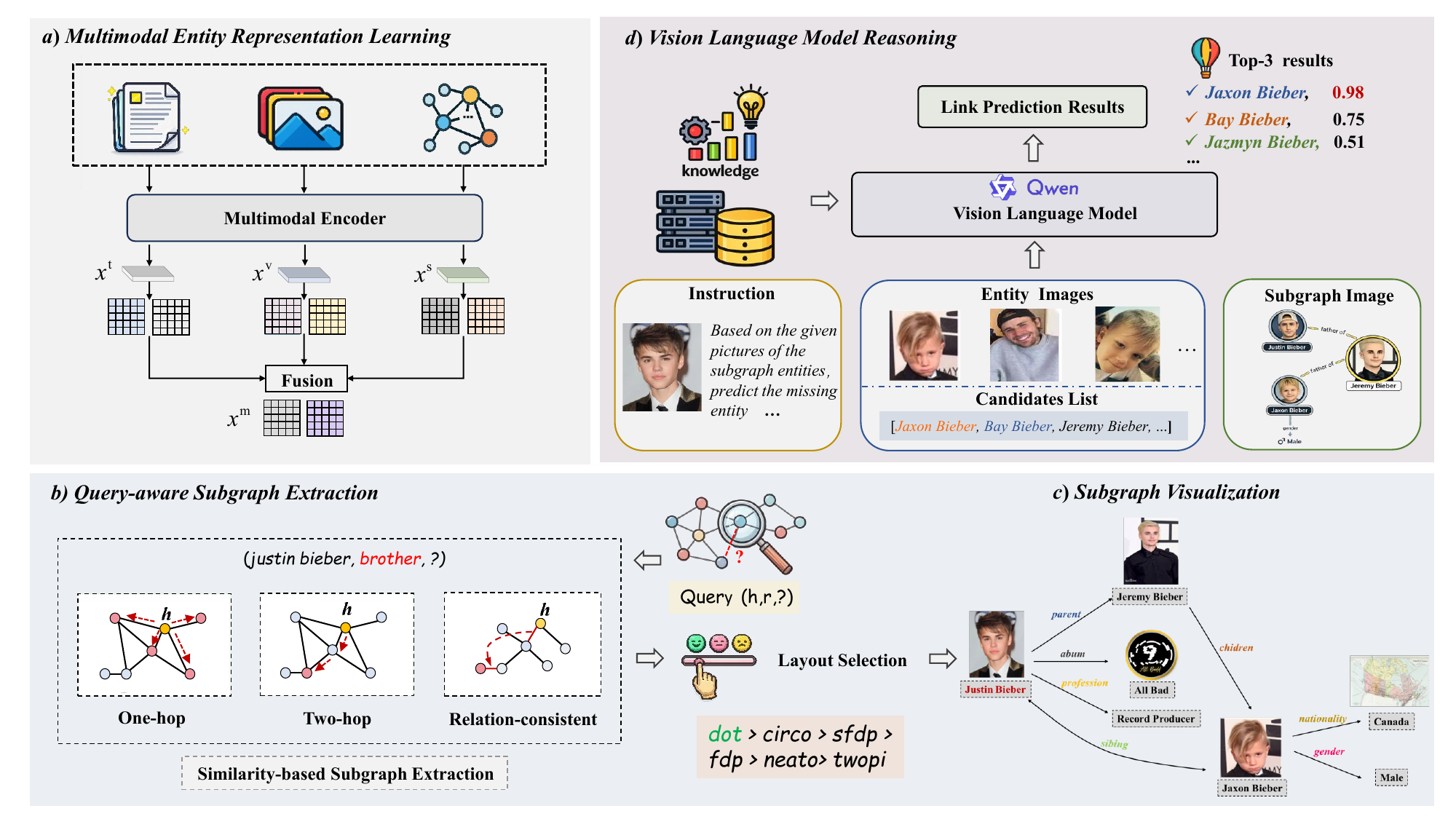}
\caption{Overview of the proposed ViSR-KGC framework. 
}
\label{fig:content2}
\end{figure*}

To predict the missing entity in MMKG, our approach leverages VLMs to integrate multimodal entity information and structural context.
The overall framework of ViSR-KGC is illustrated in Figure~\ref{fig:content2}. 
It consists of four main stages.
First, we start from \textbf{multimodal entity representation learning} using embedding-based model, to fuse textual, visual and structural information.
Second, we perform \textbf{query-aware subgraph extraction} to identify
a compact subgraph that preserves the most relevant structural
evidence for the query.
Third, we conduct \textbf{subgraph visualization} by converting the extracted subgraph into a visual representation with a graph layout strategy.
Finally, we fulfill \textbf{VLM reasoning} by jointly feeding the
visualized subgraph, entity images, as well as a structured textual prompt
containing the query, subgraph context, and candidate entities into an
VLM to infer the missing entity.

\subsection{Multimodal Entity Representation}

To obtain informative entity representations for similarity-based subgraph retrieval, we need to capture the complementarity of multimodal features through interactive fusion, while preserving modality-specific characteristics to handle potential issue of modality missing or misalignment. To this end, we adopt IMF~\cite{DBLP:conf/www/LiZXZX23} as our multimodal encoder. It provides effective joint representations across textual, visual, and structural modalities, achieving near-best prediction performance among embedding-based methods, as validated in our experiments. As the first step of the framework, it can in principle be replaced by other representation learning models.

Specifically, after training on the MMKG \(\mathcal{G}\), each entity
\(e \in \mathcal{E}\) is associated with four embeddings: a textual
embedding \(\mathbf{x}^\textrm{t}\), a visual embedding \(\mathbf{x}^\textrm{v}\), a structural embedding
\(\mathbf{x}^\textrm{s}\), and a fused multimodal embedding
\(\mathbf{x}^\textrm{m}\). That is formalized as follows.
\begin{equation}
(\mathbf{x}^\textrm{t}_i,\mathbf{x}^\textrm{v}_i,\mathbf{x}^\textrm{s}_i,\mathbf{x}^\textrm{m}_i)=f(\mathcal{G}, e),
\end{equation}
where the structural embedding is learned solely from the graph structure to highlight topological information for relational reasoning.
These representations together provide the basis for measuring the query-candidate similarity in subgraph extraction, 
supporting the subsequent multimodal reasoning process.




\subsection{Query-aware Subgraph Extraction}

Given a query triple \( (h, r, ?) \) or \( (?, r, t) \), directly
reasoning over the entire knowledge graph is infeasible for VLMs and would
introduce a large amount of irrelevant information. Therefore, we
extract a compact query-aware subgraph that preserves the most relevant structural and semantic evidence around the query, which can then be
visualized and used for downstream multimodal reasoning.

For clarity, here we present the process for tail entity prediction $(h, r, ?)$ with $h$ as the anchor entity, and head prediction can be performed with anchor $t$ similarly. 
We first define a set of candidate edges $\{(h_i,r_i,t_i)\}\subseteq \mathcal{T}$ from the training graph, composed of three categories according to their structural relationship with the query:

\begin{itemize}

\item \textbf{Relation-consistent edges} \( C_{\text{rel}} \):
edges whose relation type \( r_i \) matches the query relation \( r \),
while the head entity differs from the query entity, i.e.,
\( r_i = r \) and \( h_i \neq h \). Such edges provide additional
instances of the relational pattern.

\item \textbf{One-hop neighbor edges} \( C_{\text{1-hop}} \):
edges directly connected to the query entity \( h \), i.e.,
$h_i=h$ or $t_i=h$.
They offer the most immediate structural context.

\item \textbf{Two-hop neighbor edges} \( C_{\text{2-hop}} \):
edges connected to the one-hop neighbors of \( h \), 
but excluding those that still contain the
query entity \( h \). They capture indirect structural relations
and potential multi-hop reasoning paths.

\end{itemize}

After collecting these candidate edges, we evaluate their semantic relevance to the query using multimodal embeddings. 
For each candidate edge \( (h_i,r_i,t_i) \), we compute a relevance
score by jointly considering relation similarity and entity similarity:
\begin{equation} \label{equ:score1}
\mathrm{score}(h_i,r_i,t_i)
=
\lambda \,\mathrm{sim}(r,r_i)
+
(1-\lambda)\max\bigl(\mathrm{sim}(h,h_i), \mathrm{sim}(h,t_i)\bigr),
\end{equation}
where a hyperparameter \( \lambda \in [0,1] \) balances the two perspectives of similarity. 
The entity similarity is computed from the
four-view entity embeddings as:
\begin{equation}
\mathrm{sim}(\mathbf{x}_i, \mathbf{x}_j)
=
0.25 \times
\sum_{m \in \{\textrm{t}, \textrm{v}, \textrm{s}, \textrm{m}\}}
\cos(\mathbf{x}_i^{m}, \mathbf{x}_j^{m}).
\end{equation}
If the chosen representation learning model cannot provide all of four embeddings, e.g. when structural features are missing, the similarity can still be computed by averaging over available modalities.

To further emphasize structural proximity, we add a small bonus to those edges sharing at least one element with the query:
\begin{equation}\label{equ:score2}
\mathrm{score}^*(h_i,r_i,t_i)
=
\mathrm{score}(h_i,r_i,t_i) + \gamma \cdot \delta\bigl((h_i,r_i,t_i)\in C_{\text{1-hop}}\cup C_{\text{rel}}\bigr),
\end{equation}
where \( \gamma \) is a small positive constant, and \( \delta(\cdot) \)
equals \(1\) if the condition holds and \(0\) otherwise.

Finally, we identify the most relevant edges to construct the query subgraph $G_{\text{sub}} = (V_{\text{sub}}, E_{\text{sub}})$. 
Specifically, we select the top
\( k_{\text{rel}} \) relation-consistent edges and the top
\( k_{\text{nei}} \) neighbor edges according to their relevance
scores, merge them, remove duplicates, and truncate the merged edge set
to a maximum size \( k_{\max} \). 
This subgraph preserves local structural evidence with multimodal semantic signals, providing informative context for subsequent visualization and VLM reasoning. The procedure is detailed in Algorithm~\ref{alg:subgraph_extraction}.

\begin{algorithm}[t]
\caption{Query-aware Subgraph Extraction}
\label{alg:subgraph_extraction}
\begin{algorithmic}[1]

\Require Query triple $(h,r,?)$, training triple set $T_{\text{train}}$, hyperparameters $\lambda$, $\gamma$, $k_{\text{rel}}$, $k_{\text{nei}}$, $k_{\max}$. ($(?,r,t)$ is symmetric and omitted here) 
\Ensure Query-aware subgraph $G_{\text{sub}} = (V_{\text{sub}}, E_{\text{sub}})$.
\State Initialize candidate sets $C_{\text{rel}}$, $C_{\text{1-hop}}$, and $C_{\text{2-hop}}$ as empty sets;
\ForAll{$(h_i,r_i,t_i)\in T_{\text{train}}$}
    \If{$r_i = r \land h_i \neq h$}
        \State Add $(h_i,r_i,t_i)$ to $C_{\text{rel}}$;
    \EndIf
    \If{$h_i = h \lor t_i = h$}
        \State Add $(h_i,r_i,t_i)$ to $C_{\text{1-hop}}$;
    \EndIf
\EndFor
\State Collect one-hop neighbor entity set $V_{\text{1-hop}}$ from $C_{\text{1-hop}}$;
\ForAll{$(h_i,r_i,t_i)\in T_{\text{train}}$}
    \If{$(h_i \in V_{\text{1-hop}} \lor t_i \in V_{\text{1-hop}}) \land h_i \neq h \land t_i \neq h$}
        \State Add $(h_i,r_i,t_i)$ to $C_{\text{2-hop}}$;
    \EndIf
\EndFor
\State Initialize scored edge sets $S_{\text{rel}}$ and $S_{\text{nei}}$ as empty sets;
\ForAll{$(h_i,r_i,t_i)\in C_{\text{1-hop}} \cup C_{\text{2-hop}} \cup C_{\text{rel}}$}
    \State Compute the relevance score according to Equation \ref{equ:score2};
    \If{$(h_i,r_i,t_i)\in C_{\text{rel}}$}
        \State Add $((h_i,r_i,t_i), \mathrm{score}^*(h_i,r_i,t_i))$ to $S_{\text{rel}}$;
    \Else
        \State Add $((h_i,r_i,t_i), \mathrm{score}^*(h_i,r_i,t_i))$ to $S_{\text{nei}}$;
    \EndIf
\EndFor
\State Select top-$k_{\text{rel}}$ edges from $S_{\text{rel}}$ and top-$k_{\text{nei}}$ edges from $S_{\text{nei}}$;
\State Merge selected edges and remove duplicates to obtain $S_{\text{final}}$;
\State Truncate $S_{\text{final}}$ to at most $k_{\max}$ edges;
\State $E_{\text{sub}} \gets \{(h_i,r_i,t_i)\mid ((h_i,r_i,t_i),\mathrm{score}) \in S_{\text{final}}\}$;
\State $V_{\text{sub}} \gets \{h_i,t_i \mid (h_i,r_i,t_i)\in E_{\text{sub}}\}$;
\State \Return $G_{\text{sub}} = (V_{\text{sub}}, E_{\text{sub}})$.
\end{algorithmic}
\end{algorithm}





\subsection{Subgraph Visualization}

To enable multimodal reasoning over graph structures, the extracted subgraph is further transformed into a visual representation that can be directly handled by VLMs.
Formally, given the query-aware subgraph
\( G_{\text{sub}} \),
we apply a graph rendering function to project the graph into a
two-dimensional visual space:
\begin{equation}
I_{\text{sub}} = \mathcal{R}(G_{\text{sub}}; \ell),
\end{equation}
where \(\mathcal{R}(\cdot)\) denotes the graph rendering operation,
\(\ell\) denotes the layout strategy, and \(I_{\text{sub}}\) is the
resulting subgraph image.

In practice, this process can be implemented by various graphic
visualization tools, such as Graphviz, Matplotlib, and NetworkX.
Among them, Graphviz is particularly suitable for building large-scale
datasets as it can automatically design graph layouts with customizable
visual attributes. Following recent VLM-based work such as GITA~\cite{DBLP:conf/nips/WeiFJZZWK024}
and VisionGraph~\cite{10.5555/3692070.3693188}, we can adjust multiple
visual elements including backdrop colors, node shapes, and layout
strategies to optimize graph readability.

As GITA~\cite{DBLP:conf/nips/WeiFJZZWK024} points out, layout selection has the most significant impact on VLM
performance for graph reasoning tasks.
In this work, we investigate several commonly used layout algorithms ~\cite{10.5555/358668.358697},
including \textit{dot}, \textit{circo}, \textit{twopi}, \textit{neato},
\textit{fdp}, and \textit{sfdp}. They differ in how nodes and
edges are spatially organized, which may affect the ability of VLMs to
recognize graph topology and relational patterns. Among them, the
\textit{dot} layout arranges nodes in a hierarchical manner and reduces
edge crossings, making the resulting graph structure more readable and
interpretable. Based on the empirical comparison reported in the Supplementary Materials, we use \textit{dot} as the default layout strategy. 

\subsection{VLM Reasoning}

After constructing the visualized subgraph, we leverage VLMs to perform link prediction by jointly reasoning
over structural, visual, and textual information.

Specifically, the prompt provided to the VLM consists of two components:
\textbf{textual input} and \textbf{visual input}. The textual part \(T_{\text{prompt}}\) includes: (i) the incomplete
query triple, (ii) a textual serialization of the key triples in the
extracted query-aware subgraph, (iii) the candidate entity set
\(C=\{e_1,e_2,\dots,e_K\}\) computed by the representation learning model, and (iv) task-specific output instructions.
The visual input contains: (i) the rendered subgraph image
\(I_{\text{sub}}\), which presents the topology of the extracted
subgraph, and (ii) the entity images \(\{I_e\}_{e \in V_{\text{sub}}}\)
embodying entity-level visual semantics.

Given the multimodal prompt, VLM generates the final prediction based on the textual and visual evidence, 
with the candidate list serving as guidance rather than a hard constraint.
Formally, the prediction target can be expressed over the full entity set $\mathcal{E}$:
\begin{equation}
\hat{e}=\arg\max_{e_i \in \mathcal{E}}
P\!\left(
e_i \mid T_{\text{prompt}}, I_{\text{sub}}, \{I_e\}_{e \in V_{\text{sub}}}
 \right).
\end{equation}

In practice, ViSR-KGC employs a prompt-based generation paradigm and does not explicitly score every entity in the MMKG.
A detailed prompt template and a prompt illustration are both provided in the Supplementary Materials.

\section{Experiments}

\begin{table}[t]
\centering
\caption{Dataset statistics of the final datasets}
\label{tab:statistic}
\begin{tabular}{lccccc}
\toprule
\textbf{Dataset} & \textbf{Nodes} & \textbf{Edge Types} & \textbf{Train} & \textbf{Val} & \textbf{Test} \\
\midrule
\textbf{FB15K-237} & 14,541 & 237 & 272,115 & 17,535 & 6,102 \\
\textbf{DB15K} & 14,777 & 279 & 56,881 & 9,903 & 8,355 \\
\bottomrule
\end{tabular}
\end{table}

\begin{table*}[htbp]
\centering
\caption{Main experimental results on FB15K-237 and DB15K datasets. Some baseline results are directly copied from the corresponding papers under the same settings. The best results are highlighted in \textbf{bold}, while the second-best are \underline{underlined}.}
\label{tab:exp_results}
\small
\begin{tabular}{@{}clccccccccc@{}}
\toprule
 & & \multicolumn{4}{c}{\textbf{FB15K-237}} & \multicolumn{4}{c}{\textbf{DB15K}} \\
\cmidrule(lr){3-6} \cmidrule(l){7-10}
& & \multicolumn{2}{c}{Tail Entity Prediction} & \multicolumn{2}{c}{Head Entity Prediction} & \multicolumn{2}{c}{Tail Entity Prediction} & \multicolumn{2}{c}{Head Entity Prediction} \\
\cmidrule(lr){3-4} \cmidrule(lr){5-6} \cmidrule(lr){7-8} \cmidrule(l){9-10}
Category & Model            & Hit@1 & Hit@3 & Hit@1 & Hit@3 & Hit@1 & Hit@3 & Hit@1 & Hit@3 \\
\midrule
\multirow{2}{*}{Structure-only embedding-based} & ConvE            & 0.6919 & 0.8003 & 0.3433 & 0.4785 & 0.5274 & 0.6420 & 0.3347 & 0.4255 \\
 & ConvKB           & 0.6386 & 0.7301 & 0.2045 & 0.3367 & 0.2915 & 0.4001 & 0.2293 & 0.3181 \\
\midrule
\multirow{3}{*}{Text-enhanced embedding-based} & StAR      & 0.2660 & 0.4040 & 0.2331 & 0.3759 & 0.2254 & 0.3476 & 0.2027& 0.3011\\
& SimKGC    & 0.2520 & 0.3640 & 0.2276 & 0.3548 & 0.2379 & 0.3528 & 0.2145& 0.3093 \\
& CoLE     & 0.2940 & 0.4290 &0.2642 & 0.3987& 0.2746 & 0.3718 & 0.2354 & 0.3247\\
\midrule
\multirow{5}{*}{Multimodal embedding-based} & IMF              & 0.7399 & 0.8390 & 0.4142 & 0.5613 & 0.5794 & 0.6735 & 0.3406 & 0.4348 \\
& HGNN-IMA         & 0.7362 & 0.8243 & 0.3989 & 0.5388 & 0.5271 & 0.6214 & \underline{0.3486} & 0.4345 \\
& NativE           & 0.7147 & 0.8263 & 0.4016 & 0.5573 & 0.4823 & - & 0.2801 & - \\
& MyGO             & 0.7467 & 0.8429 & 0.4223 & 0.5679 & 0.5137 & 0.6089 & 0.3008 & 0.4126 \\
& IMVIA            & 0.7215 & 0.8152 & 0.4119 & 0.5421 & 0.5160 & 0.6092 & 0.3029 & 0.4222 \\
\midrule
\multirow{2}{*}{Unimodal LLM-based} & DIFT             & 0.3640 & 0.4680 & 0.2475 & 0.3871 & 0.3324 & 0.4276 & 0.2243 & 0.3769 \\
& SLiNT             & 0.3680 & 0.4720 & 0.2579 & 0.3924 & 0.3328 & 0.4330 & 0.2274 & 0.3812 \\
\midrule
\multirow{2}{*}{Multimodal LLM-based} & MR-MKG     & 0.7620 & 0.8543 & 0.4571 & 0.5879 & 0.6212 & 0.7015 & 0.3056 & 0.4380 \\
& MLaGA             & \underline{0.7914} & \underline{0.8733} & \underline{0.4892} & \underline{0.6177} &\underline{0.6689} &\underline{0.7723} 
& 0.3481 
& \underline{0.4673} \\
\midrule
\multirow{2}{*}{Multimodal VLM-based} & ELMM             & 0.3740 & 0.5020 & 0.2783 & 0.4229 & 0.3410 & 0.4550 & 0.2357 & 0.3977 \\
\cmidrule{2-10}
& \textbf{ViSR-KGC} & \textbf{0.8027} & \textbf{0.8836} & \textbf{0.5036} & \textbf{0.6209} & \textbf{0.6715} & \textbf{0.7802} & \textbf{0.3589} & \textbf{0.4758} \\
\bottomrule
\end{tabular}
\end{table*}

\subsection{Datasets}

We conduct experiments on two widely used multimodal knowledge graph datasets: \textbf{FB15K-237}~\cite{DBLP:conf/emnlp/ToutanovaCPPCG15} and \textbf{DB15K}~\cite{lehmann2015dbpedia}. 
FB15K-237 is a refined version of FB15K from Freebase with inverse relations removed to avoid information leakage. Its multimodal extension provides textual descriptions and images for each entity. 
DB15K is derived from DBpedia and preserves rich semantic relations, also containing textual summaries and visual information. 

To evaluate link prediction performance, we apply strict pre-processing to the test set, retaining only triples with a unique correct answer for each query $(h, r, ?)$ or $(?, r, t)$~\cite{DBLP:journals/corr/abs-2405-13640}. 
This filtering removes ambiguous queries with multiple valid answers, ensuring the evaluation can faithfully reflect the model's reasoning capability.
Detailed statistics of the filtered datasets are shown in Table~\ref{tab:statistic}.

\subsection{Baseline Models}

We compare ViSR-KGC with baselines from six categories.

\textbf{Structure-only KGC baselines.}
\textbf{ConvE}~\cite{10.5555/3504035.3504256} is a representative structural method based on convolution.
\textbf{ConvKB}~\cite{nguyen-etal-2018-novel} applies
convolution over tensorized triples for entity-relation
interactions.

\textbf{Text-enhanced KGC baselines.}
\textbf{StAR}~\cite{DBLP:conf/www/WangSLZW021} introduces textual descriptions to structral KGC through joint representation learning.
\textbf{SimKGC}~\cite{wang-etal-2022-simkgc} employs a bi-encoder and contrastive learning to enhance prediction.
\textbf{CoLE}~\cite{DBLP:conf/cikm/LiuSLH22} combines structural and textual knowledge via co-distillation with pre-trained language models.

\textbf{Multimodal embedding-based KGC baselines.}
\textbf{IMF}~\cite{DBLP:conf/www/LiZXZX23} models cross-modal
interactions through bilinear pooling and contrastive learning.
\textbf{HGNN-IMA}~\cite{DBLP:conf/ijcai/Li00L0WYW25} incorporates
cross-modal influence modeling into a heterogeneous GNN framework.
\textbf{NativE}~\cite{zhang2024native} addresses modality diversity
and imbalance issue via adaptive multimodal fusion.
\textbf{MyGO}~\cite{zhang2025tokenization} performs fine-grained
multimodal tokenization and cross-modal interaction learning.
\textbf{IMVIA}~\cite{li2025unifying} combines intra-modality
multi-view aggregation, cross-modal alignment, and relation-aware
fusion.

\textbf{LLM-based KGC baselines.}
\textbf{DIFT}~\cite{DBLP:conf/semweb/LiuTSH24} injects structural
embeddings into an LLM for candidate-aware KGC.
\textbf{SLiNT}~\cite{yang-etal-2025-slint} improves KGC via
contrastive learning and structural injection into an
LLM.

\textbf{Multimodal LLM-based reasoning baselines.}
\textbf{MR-MKG}~\cite{lee-etal-2024-multimodal} injects multimodal
knowledge into LLMs via RGAT encoding and adapter layers.
\textbf{MLaGA}~\cite{DBLP:journals/corr/abs-2506-02568} aligns text, image, and graph features through a structural encoder and adapts
LLM via instruction tuning.

\textbf{Multimodal VLM-based reasoning baselines.}
\textbf{ELMM}~\cite{huang2026elmmefficientlightweightmultimodal} introduces a lightweight VLM with visual token
compression and attention pruning for efficient multimodal reasoning.

\subsection{Evaluation Metrics}

Different from conventional score-based KGC models that emphasize full-ranking evaluation, ViSR-KGC is a prompt-based multimodal reasoning framework whose practical goal is to identify the correct entity within the top few predictions. For this reason, we adopt precision-oriented Hit@1 and Hit@3 as metrics~\cite{10.5555/2999792.2999923,DBLP:conf/kgcw/ZhangS25}.
The results are reported separately for head and tail predictions to account for varying difficulties introduced by asymmetric relations.




\subsection{Implementation Details}

ViSR-KGC employs Qwen3VL~\cite{bai2025qwen3vltechnicalreport} as the VLM backbone. 
In subgraph extraction, 
the maximum edge number is set to $k_{\text{max}}=15$ for finally extracted edges in the subgraph, $k_{\text{rel}}=10$ for same-relation edges, and $k_{\text{nei}}=10$ for neighbor edges.
The similarity weighting coefficient in Equation \ref{equ:score1} is set to $\lambda=0.5$, and the bonus weight in Equation \ref{equ:score2} is set to $\gamma=0.1$.
All experiments are conducted on a server equipped with an NVIDIA RTX 4090 GPU.

\begin{figure*}[t]
\centering
\includegraphics[width=0.8\linewidth]{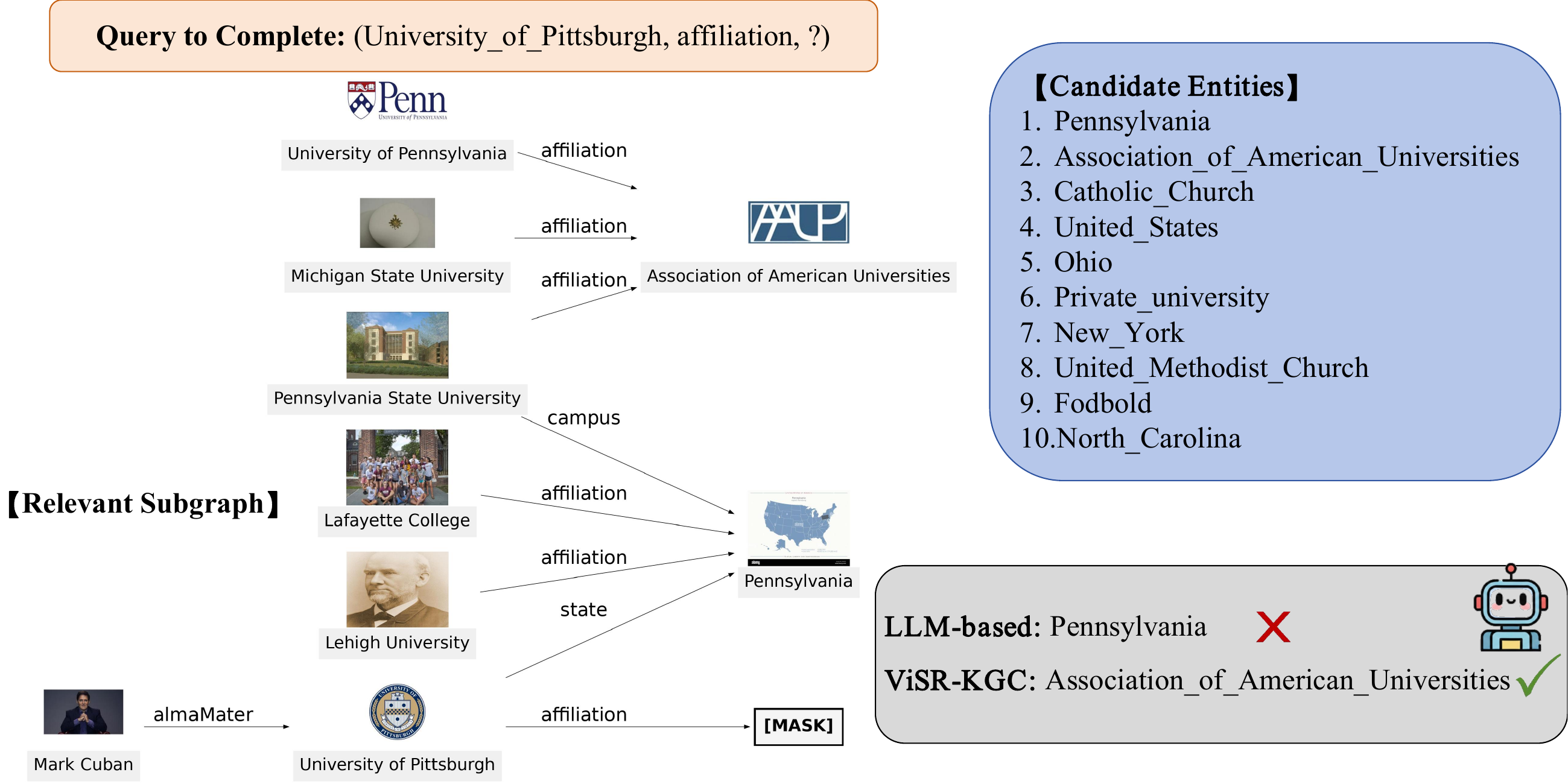}
\caption{Example of link prediction by ViSR-KGC compared to LLM-based methods.}
\label{fig:case2}
\end{figure*}

\subsection{Overall Results}

Table~\ref{tab:exp_results} reports the performance of different methods on the two datasets.
We analyze the results from several perspectives.

\textbf{(1) Traditional embedding-based methods struggle with complex reasoning.}
We can see that traditional embedding-based methods and multimodal fusion approaches achieve substantially lower performance compared to large-model-based methods.
This gap confirms that
pure representation learning, which relies mainly on graph structure and embedding spaces, struggles to capture implicit relational patterns,
particularly for low-frequency or sparsely observed relations. 
In contrast, large models benefit from strong semantic understanding and reasoning ability acquired during pretraining.
In particular, we also examine predictions on relation types that appear fewer times than the quartile in the training set, accounting for 7.6\% and 6.4\% of all edges for FB15K-237 and DB15K, respectively. In these cases, the representation learning method IMF attains Hit@1 scores of 77.14\% and 37.5\% on the two datasets, compared to 85.16\% and 48.44\% for ViSR-KGC.





More importantly, \textbf{ViSR-KGC leverages commonsense knowledge beyond candidate sets obtained from the representation learning.} Around 9.96\% and 11.24\% of correct predicted entities on FB15K-237 and DB15K respectively, fall outside the candidate set.
For example, in the query
$(?, creator, Mike\_Henry\_(voice\_actor))$,
the correct answer 
$This\_\allowbreak Is\_\allowbreak The\_\allowbreak Cleveland\_\allowbreak Show$
is not in the candidate list. However, ViSR-KGC can infer
the answer by analyzing related entities in the subgraph (e.g.,
\textit{Family Guy} and \textit{American Dad}), supplemented by its internal knowledge of television animation
production. This indicates that it performs active reasoning rather than simple pattern matching, leveraging VLMs' pretrained world knowledge and commonsense reasoning capabilities to compensate for missing information in structured graph data.


\textbf{(2) LLM-based methods lack explicit graph structure perception.}
Among LLM-based baselines that operate on textualized graph inputs, most exhibit only modest performance. Even when augmented with
multimodal information, methods like MR-MKG and MLaGA, still
process graph structure implicitly through feature propagation.
Their performance, while higher than traditional methods, remains
below our ViSR-KGC. This shows existing
LLM-based methods fail to present graph topology in a
form that can be visually perceived, limiting their ability of inferring new relations.

However, \textbf{Visualizing graph structure in ViSR-KGC facilitates more explicit structural reasoning.}
Through extracting query-aware
subgraphs and rendering them as visual representations, our approach achieves the best performance across all metrics. 
This improvement demonstrates the complementary role of multimodal information and structural visualization in the KGC task.
While textual descriptions provide semantic attributes and relational context, visual images capture appearance features and scene-level information, and visualized subgraphs explicitly reveal topological relations that are hard to infer from text alone.
As shown in Figure~\ref{fig:case2},
the query aims to identify an entity that has the
\textit{affiliation} relation with
\textit{University\_of\_Pittsburgh}.
LLM predicts \textit{Pennsylvania},
mistakenly interpreting the relationship as a geographic association.
In contrast, ViSR-KGC successfully identifies the correct
entity by inspecting the visualized subgraph structure.
Specifically, the visualization reveals a critical path
\textit{University\_of\_Pittsburgh}
$\rightarrow$
\textit{Pennsylvania}
$\rightarrow$
\textit{Pennsylvania State University}
$\rightarrow$
\textit{Association\_of\_American\_Universities},
which implicitly encodes the affiliation relationship.
This example suggests that visualized subgraph structures provide additional topology-aware signals that are difficult to recover from textual descriptions alone,
especially for difficult cases involving structural disambiguation.



\textbf{(3) Asymmetry between head and tail entity prediction.}
Another observation is that all methods perform significantly better on tail entity prediction than on head entity prediction. 
This asymmetry reflects the directional
characteristics of many relations in knowledge graphs, where head
entities are often more specific subjects (e.g. a person or organization), and tail entities are more general objects (e.g., a work or a location). Thus, predicting tail entities from head entities is usually easier than the reverse task.

\subsection{Ablation Study}

To evaluate the contribution of key components, we conduct
ablation study on the DB15K dataset.
Table~\ref{tab:ablation_db15k} reports the performance after removing or changing specific components from the full model.
The analysis below directly corresponds to key challenges.

\textbf{(1) Query-aware subgraph extraction with multi-type edges provides focused evidence (Challenge 1).}
Removing relation-consistent edges (\textbf{ViSR-KGC w/o $C_\text{rel}$}) or two-hop neighbor edges (\textbf{ViSR-KGC w/o $C_{2\text{-hop}}$}) both notably decrease the prediction accuracy, 
explaining that same-relation facts provide direct evidence for relational pattern learning,
and broader structural context also contributes useful reasoning signals. 
Interestingly, the latter variant achieves a higher Hit@3 score for head entity prediction than the full model, implying that some two-hop paths may introduce noise in harder but fault-tolerant tasks. 
Overall, extracting a query-aware subgraph with multi-type edges is essential for offering the VLM with sufficient yet focused structural evidence.

\textbf{(2) Subgraph structure with suitable layout provides critical reasoning signals (Challenge 2).}
When both the textual subgraph description and the subgraph structure image are removed (\textbf{ViSR-KGC w/o subgraph}), the model performance drops dramatically.
This indicates that subgraph structure provides essential
contextual information for understanding relationships between
entities.
Further analysis shows that the two forms of subgraph representation are complementary.
Removing the textual subgraph description (\textbf{ViSR-KGC w/o subgraph text}) results in a significant
performance decrease, suggesting that textual descriptions provide
fine-grained semantic explanations of the local graph context, which is easy to understand for VLMs.
In contrast, removing the subgraph structure image (\textbf{ViSR-KGC w/o subgraph image}) leads to a smaller
performance drop, implying that visualized topology offers
additional structural cues but is less critical than textual context.
Moreover, changing the layout strategy (\textbf{ViSR-KGC with layout-twopi/sfdp}) leads to worse accuracy, confirming that rendering the subgraph with an appropriate layout is crucial for VLMs to explicitly perceive and reason over graph topology.

\textbf{(3) Candidate entity lists guide the reasoning space of VLMs (Challenge 3).}
Removing the candidate entity list in the prompt
(\textbf{ViSR-KGC w/o candidate entities}) causes a non-negligible
performance drop.
This certifies that candidate lists endow useful guidance toward plausible answers. Without such hints, the model
has to explore a much larger solution space, which increases prediction uncertainty and would trigger VLMs' hallucination.


\textbf{(4) Multimodal input provides complementary cues (also Challenge 3).}
Removing entity images in the prompt (\textbf{ViSR-KGC w/o entity images}) modestly reduces the performance,
indicating that
although subgraph structure visualization contributes essential topological information, entity images also provide complementary entity-level visual features as semantic context. 
That demonstrates VLMs can effectively integrate diverse information sources of multiple modalities for comprehensive reasoning.


\textbf{(5) Encoder architecture significantly affects multimodal representation quality.}
We also replace the multimodal encoder with
different backbone architectures
(\textbf{ViSR-KGC with base-ConvE, with base-ConvKB}, and \textbf{with base-HGNN-IMA}).
All variants achieve
substantially lower accuracy than the full ViSR-KGC. This gap confirms that the quality of multimodal representations directly impacts the subgraph construction, and furthermore 
limits the model's ability to capture cross-modal interactions.
The chosen structure-aware multimodal encoder with four-view embeddings proves to provide a stronger representational foundation, enabling VLMs to better exploit the visualized graph context.

\begin{table}[t]
\centering
\caption{Ablation study results on DB15K.}
\label{tab:ablation_db15k}
\resizebox{\columnwidth}{!}{
\begin{tabular}{l|cc|cc}
\toprule
\multirow{2}{*}{Model} & \multicolumn{2}{c|}{Tail Entity Prediction} & \multicolumn{2}{c}{Head Entity Prediction} \\
\cmidrule(lr){2-3} \cmidrule(l){4-5}
& Hit@1 & Hit@3 & Hit@1 & Hit@3 \\
\midrule
ViSR-KGC w/o $C_\text{rel}$ & 0.6232 & 0.7275 & 0.3046 & 0.4264 \\
ViSR-KGC w/o $C_{2\text{-hop}}$ & 0.6598 & 0.7320 & 0.3265 & \textbf{0.4976} \\
ViSR-KGC w/o subgraph & 0.4545 & 0.5859 & 0.2764 & 0.3819 \\
ViSR-KGC w/o subgraph text & 0.5075 & 0.6131 & 0.2915 & 0.3970 \\
ViSR-KGC w/o subgraph image & \underline{0.6667} & \underline{0.7778} & {0.3542} & 0.4479 \\
ViSR-KGC w/o candidate entities & 0.6533 & 0.7638 & 0.3065 & 0.4271 \\
ViSR-KGC w/o entity images & 0.6566 & 0.7374 & 0.3444 & 0.4667\\
\midrule
ViSR-KGC with base-ConvE &0.6364  &0.7273  &0.2857  &0.4742 \\
ViSR-KGC with base-ConvKB &0.5361  &0.6186  &0.3263  &0.4632 \\
ViSR-KGC with base-HGNN-IMA &0.6392  &0.7216  &0.3131  &0.4545 \\
\midrule
ViSR-KGC with layout-twopi &0.6364  &0.7475  &0.3544  &0.4684 \\
ViSR-KGC with layout-sfdp &0.6429  &0.7653  &\underline{0.3571}  &0.4694 \\
\midrule
\textbf{ViSR-KGC} & \textbf{0.6715} & \textbf{0.7802} & \textbf{0.3589} & \underline{0.4758} \\
\bottomrule
\end{tabular}
}
\end{table}

\subsection{Hyperparameter Analysis}
We focus on two important hyperparameters here, and leave the analysis of others in the Supplementary Materials.

\begin{figure}[t]
\centering
\includegraphics[width=0.7\linewidth]{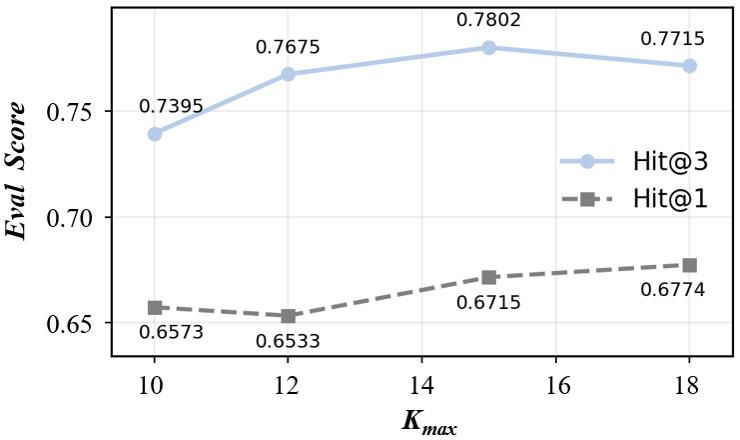}
\caption{Effect of $k_{\text{max}}$ on Hit@1 and Hit@3 on DB15K.}
\label{fig:subnumber}
\end{figure}

\textbf{Effect of $k_{\textrm{max}}$.}
We first analyze the influence of $k_{\text{max}}$, which controls the edge number of extracted subgraphs used for reasoning.
As shown in Figure~\ref{fig:subnumber}, the performance on DB15K improves as $k_{\text{max}}$ increases from 10 to 15,
indicating that larger subgraphs provide richer structural context.
When $k_{\text{max}}$ further increases to 18, Hit@3 slightly decreases,
implying that excessive subgraphs may introduce noise and increase visual complexity.
Although Hit@1 reaches the highest value at $k_{\text{max}}=18$, the result at $k_{\text{max}}=15$ is very close.
Considering both metrics, we set $k_{\text{max}}=15$ as the default value.

\begin{figure}[t]
\centering
\includegraphics[width=0.7\linewidth]{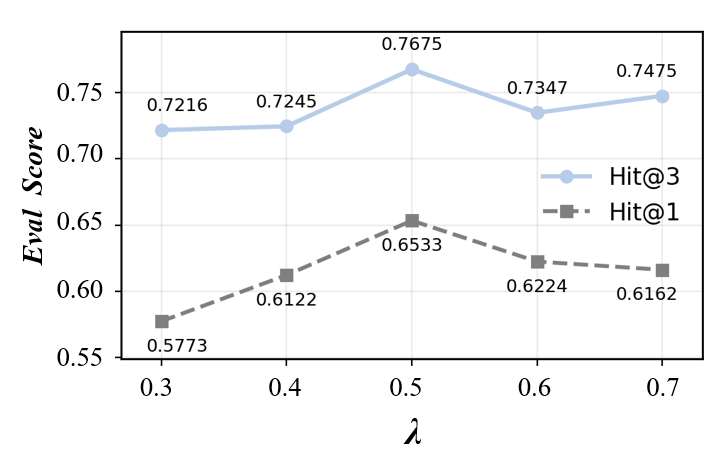}
\caption{Effect of $\lambda$ on Hit@1 and Hit@3 on DB15K.}
\label{fig:lambda}
\end{figure}

\textbf{Effect of $\lambda$.}
We further study the influence of the weighting coefficient $\lambda$, which balances relation similarity and entity similarity.
As illustrated in Figure~\ref{fig:lambda}, both Hit@1 and Hit@3 increase as $\lambda$ rises from 0.3 to 0.5, but decrease after $\lambda$ exceeds 0.5.
This suggests relation similarity is indeed beneficial, but over-emphasizing this measure would weaken the model’s ability to distinguish entities.
Therefore, $\lambda=0.5$ is a tradeoff value and used as default.

\section{Conclusion}

This paper delves into an underexplored area, multimodal knowledge graph completion (MMKGC), where the potential of LLMs and VLMs has not yet been fully leveraged. Through a systematic pipeline comprising multimodal representation learning, subgraph extraction, subgraph visualization, and prompt integration, the proposed approach, ViSR-KGC, effectively harnesses VLMs' capabilities in multimodal understanding and associative reasoning grounded in internal memory. It integrates textual, visual, and structural information, supported by a carefully chosen multimodal encoder and the layout strategy.
The synergy between representation learning and the VLM enables the capture of three levels of semantic correlations for accurate link prediction: global and local evidence within the knowledge graph, as well as implicit commonsense knowledge residing in VLMs.
Comprehensive experiments
validate the advantages of VLMs in structural reasoning and the effectiveness of each component.

Future work includes exploring more efficient subgraph extraction strategies to handle larger-scale multimodal knowledge graphs, integrating additional modalities such as audio or video to enrich multimodal semantics, and adapting the approach to other reasoning tasks on semantic graphs. Another promising direction is to adapt VLMs to MMKGC through reinforcement learning with verifiable rewards derived from link prediction correctness~\cite{liu2025visualrft,huang2026visionr1}.



\begin{acks}
This work is supported by Beijing Natural Science Foundation (L232028), CAS Project for Young Scientists in Basic Research (YSBR-040), National Natural Science Foundation of China (62373061), and National Key Laboratory of Data Space Technology and System (QZQC2026048).
\end{acks}

\appendix

\section{Dataset Filtering and Evaluation Details}

This paper uses two datasets, FB15k-237 \cite{DBLP:conf/emnlp/ToutanovaCPPCG15} and DB15k \cite{lehmann2015dbpedia}, while MMKG \cite{liu2019mmkg} provides a multimodal version for them with image attributes.
To reduce ambiguity in evaluation, we apply strict pre-processing to the
test set by retaining only queries with a unique correct answer. This
setting is intended to better match the precision-oriented objective of
our framework, where the main goal is to identify the correct entity
among the top few predictions rather than to optimize a dense global
ranking over multiple valid answers.

Specifically, for each query of the form $(h,r,?)$ or $(?,r,t)$, we
examine whether there exists exactly one gold entity in the dataset. If
multiple correct entities are associated with the same query, that query
is excluded from the final evaluation set. This design avoids the ambiguity caused
by one-to-many or many-to-one relations, ensuring that the prediction
target is well-defined and consistent with the prompting-based nature of
ViSR-KGC.

Table~\ref{tab:filter_stats} reports the dataset statistics before
and after the filtering, including the number of retained evaluation
queries.

\begin{table}[t]
    \centering
    \caption{Dataset statistics before and after unique-answer filtering.}
    \begin{tabular}{lccc}
        \toprule
       Dataset & Original & Retained & Ratio \\
        \midrule
        FB15K-237  & 20466 & 6102 & 29.81\% \\
        DB15K    & 19806 & 8355 & 42.18\% \\
        \bottomrule
    \end{tabular}
    \label{tab:filter_stats}
\end{table}


\section{Prompt Design and Templates}

This section provides additional details about the prompt construction
used in ViSR-KGC. Our prompting strategy integrates four types of input
evidence: (1) the query triple, (2) textualized local subgraph context,
(3) visual inputs including the rendered subgraph image and entity images, and
(4) a candidate entity list used as an auxiliary cue.


Specifically, given a query triple $(h,r,?)$ or $(?,r,t)$, we first construct a
query-aware local subgraph and then organize the corresponding evidence
into a multimodal prompt. The textual part includes the query, the
subgraph-derived textual context, and the candidate entity list. The
visual part includes the rendered subgraph image and the associated
entity images. It is important to note that the candidate list is used to guide the
reasoning space rather than to impose a hard constraint on decoding. The
vision-language model (VLM) makes its final prediction based on all prompt
components jointly.


Figure~\ref{fig:prompt-template} provides the generic prompt template used in our experiments, and Figure~\ref{fig:prompt-example} exhibits a concrete example.

\begin{figure}[H]
\begin{promptbox}{Prompt Template for Link Prediction Reasoning}

\textbf{Query:} \\
$(h, r, ?)$
\vspace{0.6em}

\highlightbox[respBlue]{[Relevant Subgraph]}
\vspace{0.2em}

Top-$k_{\max}$ key edges extracted from $\mathcal{G}_{\text{sub}}$, represented as triples.
\vspace{0.3em}

\highlightbox[varGreen]{[Candidate Entities]}
\vspace{0.2em}

$e_1, e_2, \ldots, e_K$
\vspace{0.3em}

\highlightbox[exYellow]{[Task Description]}
\vspace{0.2em}

You are a knowledge graph reasoning expert. Given the query, the relevant subgraph, the visual representations of the subgraph entities, and the rendered subgraph image, predict the missing entity. The candidate entity list is provided as an auxiliary cue to guide the reasoning space, but the final answer is not restricted to this list.
\vspace{0.3em}

The following visual inputs are provided:
\vspace{0.2em}

\begin{itemize}
    \item \textbf{Image $e_1$}: visual representation of entity ${entity\_name_1}$
    \item \ldots
    \item \textbf{Image $e_m$}: visual representation of entity ${entity\_name_m}$
    \item \textbf{Image $I_{\text{sub}}$}: global structural visualization of the extracted subgraph
\end{itemize}
\vspace{0.3em}

Please jointly analyze the textual and visual evidence to infer the missing entity.
\vspace{0.3em}

\highlightbox[respBlue]{[Output Format Requirements]}
\vspace{0.2em}

First line: the top-1 predicted entity name

Second line: exactly three entity names separated by commas, representing the top-3 predictions

Do not output any additional text, explanations, or formatting.

\end{promptbox}
\caption{Prompt template for link prediction reasoning.}
\label{fig:prompt-template}
\end{figure}

\begin{figure*}[!t] 
\begin{promptbox}{Illustration of the Prompt for Link Prediction Reasoning}

\textbf{Query:} \\
(Magnolia\_(film), cinematography, ?)
\vspace{0.6em}

\highlightbox[respBlue]{[Relevant Subgraph]}
\vspace{0.2em}

Top-$k_{\max}$ key edges extracted from $\mathcal{G}_{\text{sub}}$, represented as triples:
\vspace{0.2em}

Edge 1: (Punch-Drunk\_Love, cinematography, Robert\_Elswit)\\
Edge 2: (Boogie\_Nights, cinematography, Robert\_Elswit)\\
Edge 3: (There\_Will\_Be\_Blood, cinematography, Robert\_Elswit)\\
Edge 4: (Magnolia\_(film), editing, Robert\_Elswit)\\
Edge 5: (Carnage\_(2011\_film), cinematography, Alexandre\_Desplat)\\
Edge 6: (A\_Few\_Good\_Men, cinematography, Robert\_Richardson\_(cinematographer))\\
Edge 7: (Far\_from\_Heaven, cinematography, Edward\_Lachman)\\
Edge 8: (Philadelphia\_(film), cinematography, Tak\_Fujimoto)\\
Edge 9: (I\_Heart\_Huckabees, cinematography, Jon\_Brion)\\
Edge 10: (Charlie\_Wilson's\_War, cinematography, James\_Newton\_Howard)\\
Edge 11: (Gigli, cinematography, Robert\_Elswit)\\
Edge 12: (Michael\_Clayton\_(film), cinematography, Robert\_Elswit)\\
Edge 13: (Magnolia\_(film), musicComposer, Jon\_Brion)\\
Edge 14: (Good\_Night,\_and\_Good\_Luck, cinematography, Robert\_Elswit)\\
Edge 15: (Runaway\_Jury, cinematography, Robert\_Elswit)
\vspace{0.3em}

\highlightbox[varGreen]{[Candidate Entities]}
\vspace{0.2em}

Robert\_Elswit, Jon\_Brion, Roger\_Deakins, James\_Newton\_Howard, Howard\_Shore, Carter\_Burwell, Mychael\_Danna, Robert\_Richardson\_(cinematographer), Thomas\_Newman, Paul\_Thomas\_Anderson
\vspace{0.3em}

\highlightbox[exYellow]{[Task Description]}
\vspace{0.2em}

You are a knowledge graph reasoning expert. Given the query, the relevant subgraph, the visual representations of the subgraph entities, and the rendered subgraph image, predict the missing entity. The candidate entity list is provided as an auxiliary cue to guide the reasoning space, but the final answer is not restricted to this list.
\vspace{0.3em}

The following visual inputs are provided:
\vspace{0.2em}

\begin{itemize}
    \item \textbf{Image $e_1$}: visual representation of entity Punch-Drunk\_Love
    \item \textbf{Image $e_2$}: visual representation of entity Boogie\_Nights
    \item \textbf{Image $e_3$}: visual representation of entity There\_Will\_Be\_Blood
    \item \textbf{Image $e_4$}: visual representation of entity Magnolia\_(film)
    \item \textbf{Image $e_5$}: visual representation of entity Carnage\_(2011\_film)
    \item \textbf{Image $e_6$}: visual representation of entity A\_Few\_Good\_Men
    \item \textbf{Image $e_7$}: visual representation of entity Far\_from\_Heaven
    \item \textbf{Image $e_8$}: visual representation of entity Philadelphia\_(film)
    \item \textbf{Image $e_9$}: visual representation of entity I\_Heart\_Huckabees
    \item \textbf{Image $e_{10}$}: visual representation of entity Charlie\_Wilson's\_War
    \item \textbf{Image $e_{11}$}: visual representation of entity Gigli
    \item \textbf{Image $e_{12}$}: visual representation of entity Michael\_Clayton\_(film)
    \item \textbf{Image $e_{13}$}: visual representation of entity Good\_Night,\_and\_Good\_Luck
    \item \textbf{Image $e_{14}$}: visual representation of entity Runaway\_Jury
    \item \textbf{Image $e_{15}$}: visual representation of entity Robert\_Elswit
    \item \textbf{Image $e_{16}$}: visual representation of entity Jon\_Brion
    \item \textbf{Image $e_{17}$}: visual representation of entity Alexandre\_Desplat
    \item \textbf{Image $e_{18}$}: visual representation of entity Robert\_Richardson\_(cinematographer)
    \item \textbf{Image $e_{19}$}: visual representation of entity Edward\_Lachman
    \item \textbf{Image $e_{20}$}: visual representation of entity Tak\_Fujimoto
    \item \textbf{Image $e_{21}$}: visual representation of entity James\_Newton\_Howard
    \item \textbf{Image $I_{\text{sub}}$}: global structural visualization of the extracted subgraph
\end{itemize}
\vspace{0.3em}

Please jointly analyze the textual and visual evidence to infer the missing entity.
\vspace{0.3em}

\highlightbox[respBlue]{[Output Format Requirements]}
\vspace{0.2em}

First line: the top-1 predicted entity name

Second line: exactly three entity names separated by commas, representing the top-3 predictions

Do not output any additional text, explanations, or formatting

\end{promptbox}
\caption{Prompt example for link prediction reasoning.}
\label{fig:prompt-example}
\end{figure*}

\section{Layout Comparison for Subgraph Visualization}

\begin{figure*}[!htbp]
    \centering
    \begin{minipage}[b]{0.3\textwidth}
        \centering
        \includegraphics[width=\textwidth]{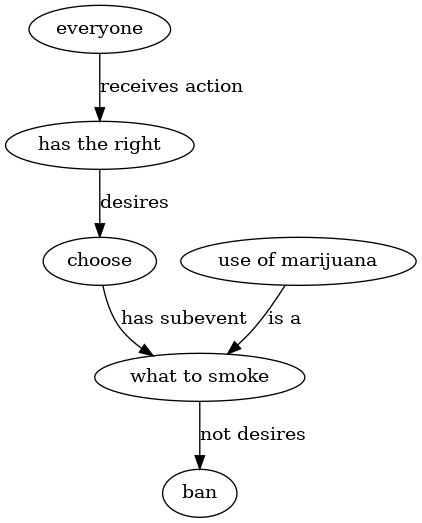}
        \caption*{(a) dot}
    \end{minipage}
    \hfill
    \begin{minipage}[b]{0.3\textwidth}
        \centering
        \includegraphics[width=\textwidth]{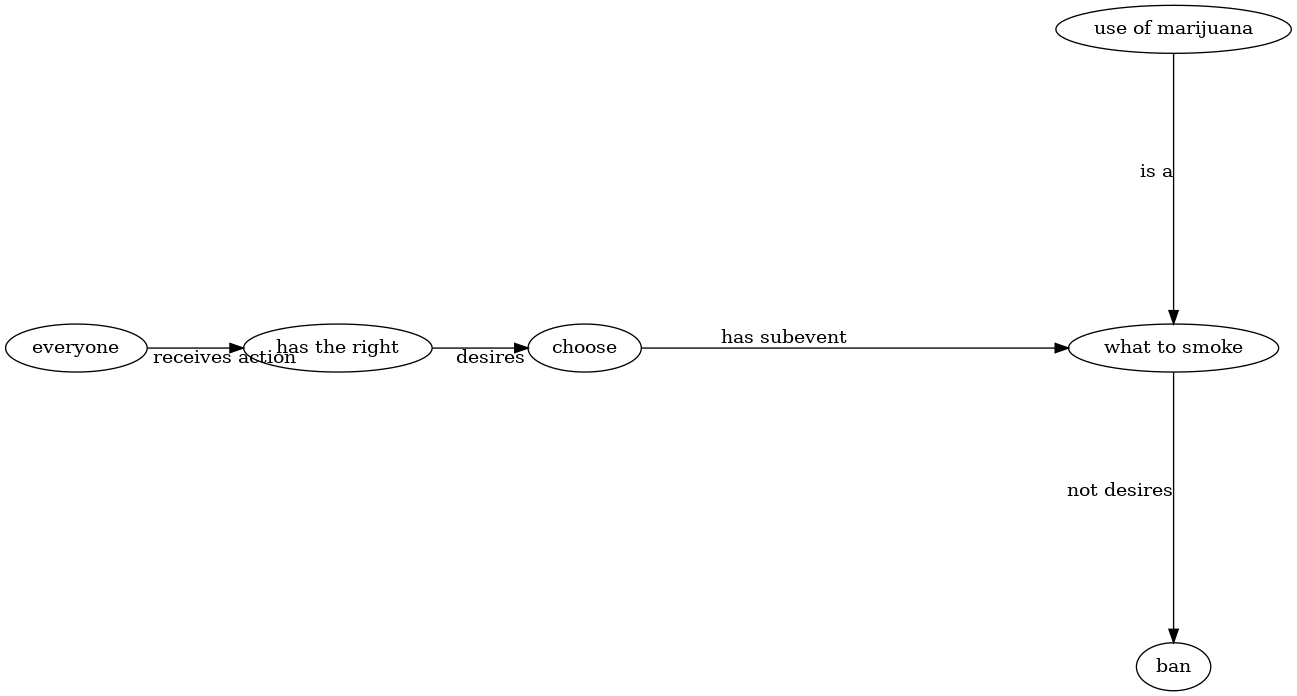}
        \caption*{(b) circo}
    \end{minipage}
    \hfill
    \begin{minipage}[b]{0.3\textwidth}
        \centering
        \includegraphics[width=\textwidth]{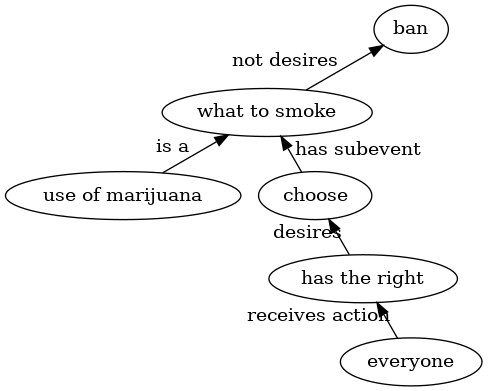}
        \caption*{(c) twopi}
    \end{minipage}
    
    \vspace{0.3cm}
    
    \begin{minipage}[b]{0.3\textwidth}
        \centering
        \includegraphics[width=\textwidth]{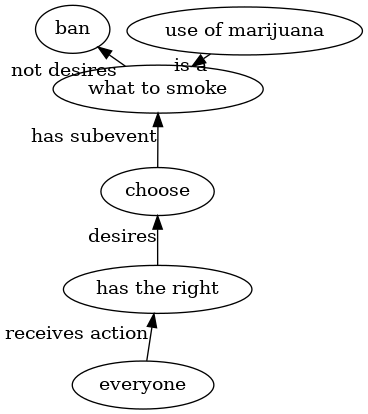}
        \caption*{(d) neato}
    \end{minipage}
    \hfill
    \begin{minipage}[b]{0.3\textwidth}
        \centering
        \includegraphics[width=\textwidth]{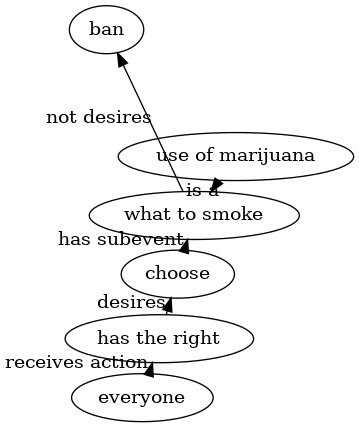}
        \caption*{(e) fdp}
    \end{minipage}
    \hfill
    \begin{minipage}[b]{0.3\textwidth}
        \centering
        \includegraphics[width=\textwidth]{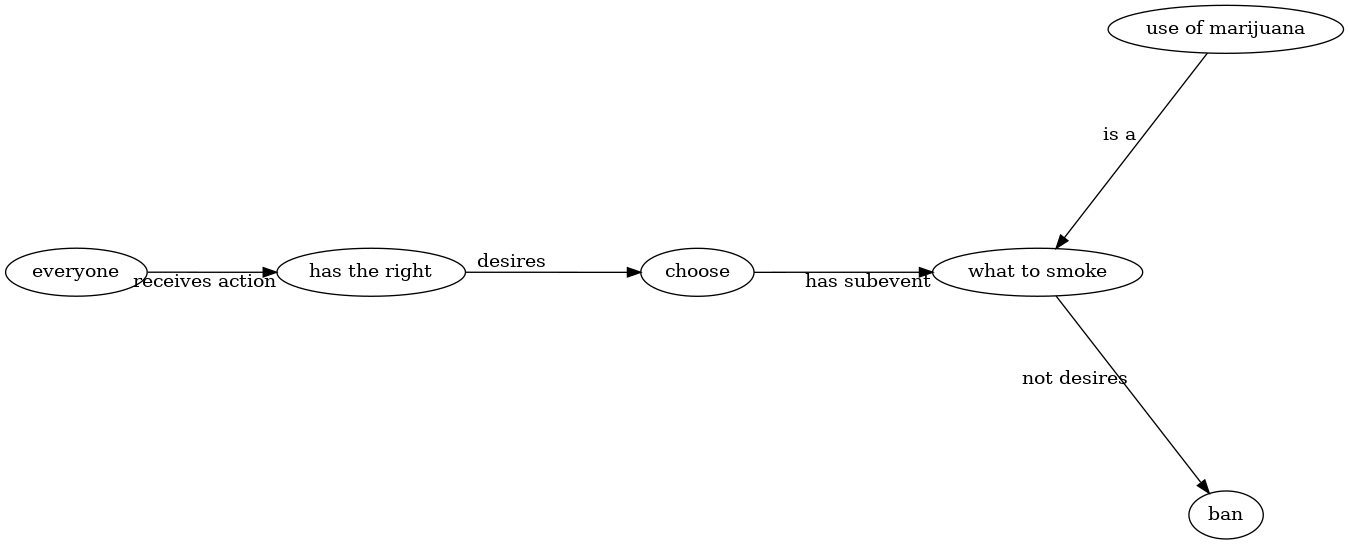}
        \caption*{(f) sfdp}
    \end{minipage}

    \caption{Examples of different layout strategies.}
    \label{fig:layout}
\end{figure*}

To examine how graph rendering affects multimodal reasoning, we compare
several graph layout strategies for subgraph visualization, including
\texttt{dot}, \texttt{circo}, \texttt{twopi}, \texttt{neato},
\texttt{fdp}, and \texttt{sfdp} provided by Graphviz \cite{10.5555/358668.358697}, which are introduced below. 
The schematic diagrams of these layouts are shown in Figure~\ref{fig:layout}.

\begin{table*}[h]
\caption{Evaluation of recognition and reasoning accuracy with different layout strategies.}
\label{tab:result2}
\centering
\resizebox{\textwidth}{!}{%
\begin{tabular}{c|ccccc|c|ccccc}
\hline
\multirow{2}{*}{\textbf{Layout}} & \multicolumn{5}{c|}{\textbf{Recognition}} & \multirow{2}{*}{} & \multicolumn{5}{c}{\textbf{Reasoning}} \\
\cline{2-6} \cline{8-12}
 & LLaVA & LLaVA-OV & InternVL2 & Qwen3VL & GPT-4o & & LLaVA & LLaVA-OV & InternVL2 & Qwen3VL & GPT-4o \\ \hline
\textbf{dot}   & \textbf{5.61} & \textbf{51.33} & \textbf{57.11} & \textbf{84.47} & \textbf{93.33} & & \textbf{22.87} & \textbf{55.34} & \textbf{65.52} & {72.10} & \textbf{95.00} \\
\textbf{circo} & 1.09 & 19.35 & 41.26 & 76.70 & 90.00 & & 7.62 & 42.38 & 58.08 & \textbf{72.26} & \textbf{95.00} \\
\textbf{twopi} & 1.25 & 5.41 & 9.03 & 11.38 & 29.27 & & 15.09 & 41.01 & 51.62 & 58.99 & 89.31 \\
\textbf{neato} & 1.67 & 11.18 & 13.80 & 22.09 & 31.25 & & 19.36 & 48.32 & 59.98 & 60.92 & 92.21 \\
\textbf{fdp}   & 1.10 & 12.62 & 15.54 & 28.48 & 23.91 & & 18.45 & 40.55 & 55.52 & 60.37 & 86.67 \\
\textbf{sfdp}  & 0.17 & 15.00 & 31.15 & 53.44 & 68.25 & & 3.20 & 37.35 & {64.32} & 71.34 & 94.12 \\ \hline
\end{tabular}%
}
\end{table*}
\begin{itemize}
\item \textbf{dot} employs a hierarchical structure, where nodes are arranged in ranks and edges are oriented in a uniform direction (top-to-bottom or left-to-right) to minimize edge crossings.
\item \textbf{circo} adopts a circular or elliptical arrangement, making it suitable for graphs with cyclic structures or strong symmetry.
\item \textbf{twopi} is a radial layout that places nodes on concentric circles around a central node. However, it tends to produce edge overlaps for non-radial structures.
\item \textbf{neato} is a reasonable default tool to use for undirected graphs that are not too large (less than 100 nodes), when you do not know anything else about the graph.
\item \textbf{fdp} implements the Fruchterman-Reingold heuristic including a multi-grid solver that handles larger graphs and clustered undirected graphs.
\item \textbf{sfdp} is a fast, multi-level, force-directed algorithm that efficiently layouts large graphs, outlined in ``Efficient and High Quality Force-Directed Graph Drawing''.
\end{itemize}

To make comparison among these layout strategies, we use the ExplaGraphs Dataset~\cite{he2024gretriever} and vary the layout style of each knowledge graph while keeping all other settings fixed. Before starting the evaluation experiments, we first conduct manual filtering to remove images that are visually unclear to the human eye, ensuring that all characters are unobstructed and the triple content remains consistent across layouts, with only the visual format differing.

Table~\ref{tab:result2} shows that there are significant differences in recognition performance across different layouts. For the LLaVA-OV, InternVL2, and Qwen3VL models, the ranking of recognition performance across layouts is consistent: \textbf{dot > circo > sfdp > fdp > neato > twopi}. For GPT-4o, which has stronger recognition capabilities, the recognition accuracy on certain layouts (e.g., twopi, neato, and fdp) is relatively lower. The LLaVA model achieves generally low triple recognition accuracy across all layouts, with accuracy reaching only about 1\% for layouts other than dot.

Based on these empirical comparisons, we select \emph{dot} as the layout strategy of the Multimodal knowledge graph completion (MMKGC) task. 
In the main paper, we present ablation results for the twopi and sfdp layouts. Here, we provide the complete ablation results for all layout strategies. The goal is to identify and verify a layout that produces visually interpretable graph structures for the VLMs.

Table~\ref{tab:layout_comp} reports the quantitative results on DB15K and FB15k-237 under different layout strategies. Among them, dot achieves the best overall performance and is therefore confirmed as the default layout in our framework. The overall trend is consistent with the layout ranking shown in Table ~\ref{tab:result2}. A plausible explanation is that dot tends to produce a clearer hierarchical organization centered around the query structure, making local connectivity patterns and relational dependencies easier for the VLM to interpret. In contrast, radial or force-directed layouts may scatter semantically related nodes, or obscure the visibility of relation chains.

\begin{table}[t]
    \centering
    \caption{Performance comparison under different graph layout strategies on DB15K and FB15K-237.}
    \setlength{\tabcolsep}{5pt}
    \renewcommand{\arraystretch}{0.95}
    \begin{tabular}{l|cc|cc}
        \toprule
        \multirow{2}{*}{Layout} 
        & \multicolumn{2}{c|}{DB15K} 
        & \multicolumn{2}{c}{FB15K-237} \\
        \cmidrule(lr){2-3} \cmidrule(l){4-5}
        & Hit@1 & Hit@3 & Hit@1 & Hit@3 \\
        \midrule
        \texttt{dot}   & \textbf{0.6715} & \textbf{0.7802} & \textbf{0.8027} & \textbf{0.8836} \\
        \texttt{circo} & 0.6512 & 0.7611 & 0.7815 & 0.8689 \\
        \texttt{sfdp}  & 0.6429 & 0.7653 & 0.7532 & 0.8406 \\
        \texttt{fdp}   & 0.6391 & 0.7526 & 0.7487 & 0.8351 \\
        \texttt{neato} & 0.6380 & 0.7504 & 0.7459 & 0.8324 \\
        \texttt{twopi} & 0.6364 & 0.7475 & 0.7460 & 0.8335 \\
        \bottomrule
    \end{tabular}
    \label{tab:layout_comp}
\end{table}



\section{Generalization Experiments on Additional Datasets and VLMs}

We conduct the evaluation of ViSR-KGC on additional entity prediction datasets, including MKG-W and MKG-Y, to test the generalization beyond DB15K and FB15K-237. Moreover, we test additional VLM backbones such as LLaVA-OV and InternVL2 on MKG-W, to verify that the complementary gains from visual-textual structural organization are consistent across various VLM models.
Table \ref{tab:extended_dataset_VLM} exhibits ViSR-KGC performance in these settings.

\begin{table}[t]
    \centering
    \caption{Generalization results on additional datasets and VLMs.}
    \label{tab:extended_dataset_VLM}

\small
\setlength{\tabcolsep}{3pt}
\renewcommand{\arraystretch}{0.90}
\begin{tabular}{l l c c}
\toprule
Model & Setting & Hits@1 & Hits@3 \\
\midrule

Candidate-only VLM & MKG-W / Qwen3VL & 0.2811 & 0.3714 \\
Textualized-subgraph-only VLM & MKG-W / Qwen3VL & 0.3230 & 0.4099\\
\textbf{ViSR-KGC} & MKG-W / Qwen3VL & \textbf{0.3723} & \textbf{0.4618} \\
\midrule

Candidate-only VLM & MKG-W / LLaVA-OV & 0.2675 & 0.3580 \\
\textbf{ViSR-KGC} & MKG-W / LLaVA-OV & \textbf{0.3516} & \textbf{0.4387} \\
\midrule

Candidate-only VLM & MKG-W / InternVL2 & 0.2984 & 0.3791 \\
\textbf{ViSR-KGC} & MKG-W / InternVL2 & \textbf{0.3785} & \textbf{0.4688} \\
\midrule

Candidate-only VLM & MKG-Y / Qwen3VL & 0.3025 & 0.3824 \\
Textualized-subgraph-only VLM & MKG-Y / Qwen3VL & 0.3412 & 0.4011 \\
\textbf{ViSR-KGC} & MKG-Y / Qwen3VL & \textbf{0.3852} & \textbf{0.4355} \\
\midrule

Candidate-only VLM & DB15K / LLaVA-OV & 0.6010 & 0.7234 \\
\textbf{ViSR-KGC} & DB15K / LLaVA-OV & \textbf{0.6286} & \textbf{0.7391} \\
Candidate-only VLM & DB15K / InternVL2 & 0.6247 & 0.7455 \\
\textbf{ViSR-KGC} & DB15K / InternVL2 & \textbf{0.6492} & \textbf{0.7587} \\

\bottomrule
\end{tabular}
\end{table}

These results show that ViSR-KGC consistently works across datasets and VLM backbones, demonstrating its robustness and generalization ability.

\section{Extended Ablation Results}

This section provides additional ablation results to complement the analysis of component roles in the main paper. We further examine the effects of
different prompt components, encoder backbones, and visualization
settings on the other dataset FB15K-237. Table~\ref{tab:extended_ablation} summarizes the
extended ablation results.

\begin{table}[t]
    \centering
    \caption{Extended ablation results on FB15K-237.}
    \label{tab:extended_ablation}
    \resizebox{\columnwidth}{!}{
    \begin{tabular}{l|cc|cc}
        \toprule
        \multirow{2}{*}{Model} & \multicolumn{2}{c|}{Tail Entity Prediction} & \multicolumn{2}{c}{Head Entity Prediction} \\
        \cmidrule(lr){2-3} \cmidrule(l){4-5}
        & Hit@1 & Hit@3 & Hit@1 & Hit@3 \\
        \midrule
        ViSR-KGC w/o $C_\text{rel}$ & 0.7318 & 0.8392 & 0.4508 &0.6011 \\
        ViSR-KGC w/o $C_{2\text{-hop}}$ & 0.7503 & 0.8465 & 0.4613 & 0.6085 \\
         ViSR-KGC w/o subgraph & 0.5979 & 0.7243 & 0.3469 & 0.4512 \\
        ViSR-KGC w/o subgraph text & 0.6813 & 0.8014 & 0.3872 & 0.5126 \\
        ViSR-KGC w/o subgraph image & 0.7938 & 0.8745 &0.4899 & 0.6207 \\
        ViSR-KGC w/o entity images & 0.7874 & 0.8681 & 0.4805 & 0.6159 \\
        ViSR-KGC w/o candidate entities & 0.7811 & 0.8634 & 0.4793 & 0.6107 \\
        \midrule
        ViSR-KGC with ConvE-base & 0.6361 & 0.7592 & 0.4391 & 0.5763\\
        ViSR-KGC with ConvKB-base & 0.6324 & 0.7553 & 0.4386 & 0.5729 \\
        ViSR-KGC with HGNN-IMA-base & 0.7402 & 0.8311 & 0.4465& 0.5818 \\
        \midrule
        ViSR-KGC with layout-twopi & 0.7460 & 0.8335 & 0.4658& 0.6042\\
        ViSR-KGC with layout-sfdp & 0.7532 & 0.8406 & 0.4707 & 0.6083 \\
        \midrule
        \textbf{ViSR-KGC} & \textbf{0.8027} & \textbf{0.8836} & \textbf{0.5036} & \textbf{0.6209} \\
        \bottomrule
    \end{tabular}
    }
\end{table}

The overall observations remain consistent with those reported on DB15K in the
main paper. Removing the textualized local subgraph causes the most
substantial degradation among the prompt-level variants, which confirms
that textualized structural context provides the dominant reasoning
signal. In contrast, removing the rendered subgraph image leads to a
much smaller drop, suggesting that visualized topology mainly offers
complementary structural cues. Candidate entities and entity images are
also beneficial, but their contributions are secondary compared with
textualized subgraph evidence.

Moreover, replacing the default encoder or layout with alternative choices also
reduces performance, which further supports the design decisions made in
the main paper. Overall, these additional results confirm that the
observed trends are robust across datasets and configurations.

\section{Extended Hyperparameter Analysis}

This section provides additional hyperparameter analysis to complement
the results reported in the main paper. We inspect the effects of
$\gamma$, $k_{\mathrm{rel}}$, $k_{\mathrm{nei}}$, and $K$ on DB15K, and
further provide supplementary validation of the two key
hyperparameters studied in the main paper, namely $k_{\max}$ and
$\lambda$, on the other dataset FB15K-237.

\subsection{Effect of $\gamma$}

The hyperparameter $\gamma$ controls the bonus weight assigned to edges in
$C_{1\text{-hop}} \cup C_{\mathrm{rel}}$, sharing at least one element with the query. We vary $\gamma$ over a range
of values and report the resulting performance on DB15K in
Table~\ref{tab:gamma_analysis}.

\begin{table}[t]
    \centering
    \caption{Effect of the bonus weight $\gamma$ on DB15K.}
    \setlength{\tabcolsep}{6pt}
    \renewcommand{\arraystretch}{0.95}
    \begin{tabular}{lcc}
        \toprule
        $\gamma$ & Hit@1 & Hit@3 \\
        \midrule
        0.00 & 0.6548 & 0.7641 \\
        0.05 & 0.6661 & 0.7756 \\
        0.10 & \textbf{0.6715} & \textbf{0.7802} \\
        0.15 & 0.6684 & 0.7773 \\
        0.20 & 0.6617 & 0.7704 \\
        \bottomrule
    \end{tabular}
    \label{tab:gamma_analysis}
\end{table}

Very small values of $\gamma$ do not sufficiently emphasize
structurally proximal edges, while too large values may overweight
such edges and introduce scoring bias. The best performance is achieved
at $\gamma=0.10$, which suggests that a modest bonus helps the model
retain useful local structure without overwhelming the semantic
relevance signals.

\subsection{Effect of $k_{\mathrm{rel}}$ and $k_{\mathrm{nei}}$}

We further analyze the effect of the numbers of retained
relation-consistent edges and neighborhood-based edges, denoted by
$k_{\mathrm{rel}}$ and $k_{\mathrm{nei}}$, respectively. The results on
DB15K are shown in Table~\ref{tab:k_analysis}.

\begin{table}[t]
    \centering
    \caption{Effect of maximum edge numbers $k_{\mathrm{rel}}$ and $k_{\mathrm{nei}}$ on DB15K.}
    \setlength{\tabcolsep}{5pt}
    \renewcommand{\arraystretch}{0.95}
    \begin{tabular}{cccc}
        \toprule
        $k_{\mathrm{rel}}$ & $k_{\mathrm{nei}}$ & Hit@1 & Hit@3 \\
        \midrule
        6  & 6  & 0.6412 & 0.7486 \\
        8  & 8  & 0.6587 & 0.7684 \\
        10 & 10 & \textbf{0.6715} & \textbf{0.7802} \\
        12 & 10 & 0.6678 & 0.7761 \\
        10 & 12 & 0.6642 & 0.7725 \\
        12 & 12 & 0.6606 & 0.7698 \\
        \bottomrule
    \end{tabular}
    \label{tab:k_analysis}
\end{table}

When these values are too small, the extracted subgraph may fail to
provide sufficient local evidence for reasoning. In contrast, large
values possibly introduce noisy or weakly related edges, which reduce prompt
quality and thus make the visualized subgraph more cluttered. The best
performance is achieved with $k_{\mathrm{rel}}=10$ and
$k_{\mathrm{nei}}=10$, indicating that moderate subgraph coverage
offers the best trade-off between evidence completeness and noise
control.

\subsection{Effect of $K$}

Finally, we study the effect of the candidate set size $K$ on DB15K.
The results are shown in Table~\ref{tab:K_analysis}.

\begin{table}[t]
    \centering
    \caption{Effect of the candidate set size $K$ on DB15K.}
    \setlength{\tabcolsep}{6pt}
    \renewcommand{\arraystretch}{0.95}
    \begin{tabular}{lcc}
        \toprule
        $K$ & Hit@1 & Hit@3 \\
        \midrule
        5  & 0.6483 & 0.7527 \\
        10 & 0.6649 & 0.7741 \\
        15 & \textbf{0.6715} & \textbf{0.7802} \\
        20 & 0.6592 & 0.7684 \\
        25 & 0.6514 & 0.7610 \\
        \bottomrule
    \end{tabular}
    \label{tab:K_analysis}
\end{table}

A smaller candidate set narrows the reasoning space but may hurt
retrieval recall, while a larger candidate set provides broader
coverage yet increases the difficulty of multimodal disambiguation. The
results show that ViSR-KGC performs best with an intermediate value of
$K$, which provides a suitable balance between recall and reasoning
difficulty.

\subsection{Effect of key $k_{\max}$ and $\lambda$ on FB15K-237}

To examine whether the hyperparameter choices are dataset-specific, we
further analyze the two key hyperparameters studied in the main paper,
namely $k_{\max}$ and $\lambda$, on FB15K-237.

\begin{table}[t]
    \centering
    \caption{Effect of the maximum final edge number $k_{\max}$ on FB15K-237.}
    \setlength{\tabcolsep}{6pt}
    \renewcommand{\arraystretch}{0.95}
    \begin{tabular}{lcc}
        \toprule
        $k_{\max}$ & Hit@1 & Hit@3 \\
        \midrule
        10 & 0.7862 & 0.8714 \\
        12 & 0.7945 & 0.8786 \\
        15 & \textbf{0.8027} & \textbf{0.8836} \\
        18 & 0.7981 & 0.8802 \\
        \bottomrule
    \end{tabular}
    \label{tab:kmax_fb}
\end{table}

\begin{table}[t]
    \centering
    \caption{Effect of the weighting coefficient $\lambda$ on FB15K-237.}
    \setlength{\tabcolsep}{6pt}
    \renewcommand{\arraystretch}{0.95}
    \begin{tabular}{lcc}
        \toprule
        $\lambda$ & Hit@1 & Hit@3 \\
        \midrule
        0.3 & 0.7643 & 0.8521 \\
        0.4 & 0.7816 & 0.8667 \\
        0.5 & \textbf{0.8027} & \textbf{0.8836} \\
        0.6 & 0.7914 & 0.8742 \\
        0.7 & 0.7868 & 0.8693 \\
        \bottomrule
    \end{tabular}
    \label{tab:lambda_fb}
\end{table}

The trends on FB15K-237 are consistent with those observed on DB15K.
Increasing $k_{\max}$ initially improves performance by providing richer
structural context, while larger values introduce more irrelevant edges
and visual clutter, leading to slight degradation. Similarly,
performance first increases with $\lambda$, indicating the importance of
relation similarity in subgraph extraction, and then decreases when
$\lambda$ becomes too large, suggesting that overemphasizing relation
similarity weakens entity discrimination.

Overall, the hyperparameter analysis shows that moderate settings
consistently outperform overly small or overly large values.
The consistent trends across DB15K and FB15K-237 further certify that the hyperparameters are not dataset-sensitive, and ViSR-KGC is reasonably robust to hyperparameter choices.

\section{Additional Case Studies}

To further illustrate the behavior of ViSR-KGC, we provide several
intuitive examples beyond the case study shown in the main paper.
Instead of presenting full visual screenshots, we summarize each case in
a concise textual form, highlighting the query, the model behavior, and
the key reason why the full framework succeeds.

\subsection{Case 1: Visualized Topology Corrects a Text-Only Error}

\textbf{Query.} $(\texttt{University\_of\_Pittsburgh}, \texttt{affiliation}, ?)$

\textbf{Text-only prediction.} ``Pennsylvania''

\textbf{Full ViSR-KGC prediction.} ``Association of American Universities''

\textbf{Gold answer.} ``Association of American Universities''

\textbf{Explanation.} In this case, the textual prompt alone leads the
model to prefer a geographically related entity, because
``Pennsylvania'' is strongly associated with
\texttt{University\_of\_Pittsburgh} at the surface semantic level.
However, the rendered subgraph makes the structural path
\texttt{University\_of\_Pittsburgh} $\rightarrow$
``Pennsylvania'' $\rightarrow$
``Association of American Universities''
more explicit, which helps the model distinguish geographical proximity
from institutional affiliation. This example suggests that visualized
topology can provide supplementary structural cues that are not always
fully captured by textual descriptions alone.

\subsection{Case 2: Correct Prediction Outside the Candidate Set}

\textbf{Query.} $(?, \texttt{creator}, \texttt{Mike\_Henry\_(voice\_actor)})$

\textbf{Candidate entities.} The candidate list does not contain:
``This Is the Cleveland Show''.

\textbf{Full ViSR-KGC prediction.}
``This Is the Cleveland Show''

\textbf{Gold answer.}
``This Is the Cleveland Show''

\textbf{Explanation.} Although the correct answer is not included in the
candidate list, the model is still able to recover it by jointly using
the local subgraph, multimodal entity evidence, and its internal
knowledge. In particular, nearby entities such as
``Family Guy'' and ``American Dad!''
provide strong contextual evidence about animated television production
and voice-actor relations. This case highlights that the candidate set
serves as a guidance cue rather than a hard constraint, and the
model is able to go beyond direct candidate matching when sufficient multimodal
evidence is available.

\subsection{Case 3: A Difficult Head Prediction}

\textbf{Query.} $(?, \texttt{language}, \texttt{English\_language})$

\textbf{Candidate entities.} The candidate list contains multiple
plausible entities that are all semantically related to
\texttt{English\_language}, such as countries, universities, books, and
films.

\textbf{Text-only prediction.} ``United States''

\textbf{Full ViSR-KGC prediction.} ``Harry Potter and the Philosopher's Stone''

\textbf{Gold answer.} ``Harry Potter and the Philosopher's Stone''

\textbf{Explanation.} This example illustrates why head prediction is
generally more difficult than tail prediction. Given the tail entity
\texttt{English\_language}, many entities may appear superficially
compatible with the relation \texttt{language}. For example,
``United States'' is strongly associated with English at the level
of common knowledge, which makes it a plausible but overly generic
prediction for a text-only model. However, the actual answer is a
specific work whose language attribute is English.

The full ViSR-KGC framework succeeds because it does not rely only on
surface semantic association. Instead, it jointly considers the
query-aware local subgraph, the textualized structural context, and the
visualized topology. In the local subgraph, the correct entity is
supported by neighboring facts that are more specific to creative works,
such as related entities of type \texttt{book}, \texttt{author}, or
\texttt{publication}. These structural cues help the model infer that
the missing head entity should be an individual work rather than a
country or organization.

This case shows that head prediction requires stronger structural
disambiguation than tail prediction, because the candidate space is
broader and semantically more diverse. It also highlights that the full
ViSR-KGC framework can exploit query-aware local evidence to distinguish
between globally related but structurally mismatched entities and the
correct answer.
\subsection{Case 4: A Zero-shot Relation Prediction}
\textbf{Query.} (?, \text{portrayer}, \text{Edgar Rice Burroughs})

\textbf{IMF prediction.} ``Island''

\textbf{Full ViSR-KGC prediction.} ``Tarzan''

\textbf{Gold answer.} ``Tarzan''

\textbf{Explanation.}
In this case, representation learning models exhibit weak zero-shot performance as the ``portrayer'' relation is unseen during training, leading to irrelevant top-$K$ results such as Island and Economist. Conversely, VISK-KGC circumvents this limitation by integrating the prior knowledge embedded in LLMs with subgraph reasoning. Specifically, by extracting the neighboring relation $(\text{Tarzan}, \text{creator}, \text{Edgar Rice Burroughs})$, VISK-KGC can logically deduce the missing entity Tarzan, underscoring its superiority in handling long-tail or even unseen relations.

\section{Complexity and Efficiency Analysis}

In addition to prediction performance, we further analyze the
computational cost of {ViSR-KGC}. Compared with conventional
embedding-based KGC models, the proposed framework introduces extra
overhead from query-aware subgraph extraction, graph rendering, and
VLM-based multimodal inference. Among these components, the dominant
cost comes from the final vision-language reasoning stage, while the
preceding stages mainly serve as lightweight or moderate
pre-processing.

\subsection{Complexity of Subgraph Construction}

Given a query triple, {ViSR-KGC} first retrieves candidate edges
from the training graph and partitions them into relation-consistent
edges, one-hop neighbors, and two-hop neighbors. Let
$|E_{\mathrm{train}}|$ denote the number of training edges. In the
worst case, identifying these candidate edge sets requires scanning the
training graph once, leading to a complexity of
$\mathcal{O}(|E_{\mathrm{train}}|)$.

For each retained candidate edge $(h_i, r_i, t_i)$, the model computes
a relevance score based on relation similarity and multimodal entity
similarity. Let $N_c$ denote the number of candidate edges to be
scored. This stage requires $\mathcal{O}(N_c)$ operations, and sorting
all candidates takes $\mathcal{O}(N_c \log N_c)$. Since only the
top-$k_{\mathrm{rel}}$ and top-$k_{\mathrm{nei}}$ edges are
finally retained, the practical subgraph size remains small. As a
result, the actual subgraph construction cost is much lower than the
worst-case bound and is dominated by operations on a compact local
candidate pool rather than the full MMKG.

\subsection{Complexity of Graph Rendering}

After subgraph extraction, the selected local graph is rendered into an
image using Graphviz. Because the final subgraph is truncated to at
most $k_{\max}$ edges, the rendering cost depends only on the size of a
small local graph rather than the full knowledge graph. Let
$\mathcal{R}(k_{\max})$ denote the rendering cost for a graph with at
most $k_{\max}$ edges. Although Graphviz layout algorithms do not admit
a single simple closed-form expression, this term is practically
bounded by the small subgraph size.

More importantly, graph rendering is a one-shot deterministic
pre-processing step on a compact graph, without large-scale neural
computation. Therefore, even though it introduces additional overhead,
its cost is typically much smaller than the subsequent multimodal VLM
inference stage.

\subsection{Complexity of Multimodal Inference}

The final prediction is generated by a VLM conditioned on the textual
prompt, the rendered subgraph image, and entity images. Let $L$ denote
the textual prompt length and $M$ denote the number of visual inputs.
Then, the cost of this stage can be expressed  as
$\mathcal{C}_{\mathrm{VLM}}(L, M)$, which includes textual encoding,
visual token processing, cross-modal interaction, and auto-regressive
decoding.

Unlike subgraph extraction and rendering, this stage requires
large-scale neural computation in a high-capacity multimodal model. In
particular, it involves: (1) encoding long textual prompts, (2)
processing one or multiple images into visual tokens, (3) performing
cross-modal reasoning between text and visual inputs, and (4)
generating the final prediction in an auto-regressive manner. 


\subsection{Overall Complexity and Practical Latency}

Based on the analysis above, the overall per-query complexity of {ViSR-KGC} can be summarized
as
\[
\mathcal{O}(|E_{\mathrm{train}}|) + \mathcal{O}(N_c \log N_c)
+ \mathcal{O}(\mathcal{R}(k_{\max}))
+ \mathcal{O}(\mathcal{C}_{\mathrm{VLM}}(L, M)).
\]
Among these terms, the first two correspond to local graph filtering
and candidate ranking, the third is related to rendering a compact
subgraph, and the last term involves large-model multimodal
reasoning.

To connect the complexity view with a more intuitive time-based
interpretation, let the average running times of subgraph extraction,
edge scoring/ranking, graph rendering, and VLM inference for a single
query be denoted by $t_{\mathrm{ext}}$,
$t_{\mathrm{rank}}$, $t_{\mathrm{render}}$, and
$t_{\mathrm{vlm}}$, respectively. Then, the total inference time can be
written as
\[
T_{\mathrm{total}}
=
t_{\mathrm{ext}} + t_{\mathrm{rank}} + t_{\mathrm{render}}
+ t_{\mathrm{vlm}}.
\]

In practice, the relative trend typically satisfies
\[
t_{\mathrm{vlm}} \gg t_{\mathrm{render}} \geq
t_{\mathrm{rank}} \geq t_{\mathrm{ext}},
\]
because the first three stages only operate on a compact local
subgraph, while the last stage requires full forward computation and
decoding in a multimodal large model. Therefore, the main trade-off of
{ViSR-KGC} is reflected in the improved multimodal reasoning capability versus the
higher inference latency.

Table~\ref{tab:efficiency_summary} summarizes the relative cost of each
stage, and Table~\ref{tab:latency_estimate} provides a representative
per-query latency estimation. Although the exact values vary with the
chosen VLM, hardware platform, prompt length, and number of images, the
overall trend is stable: VLM inference dominates the end-to-end
latency, whereas subgraph construction and graph rendering contribute
only a comparatively small fraction of the total cost.

\begin{table}[H]
    \centering
    \caption{Qualitative summary of the relative computational cost of
    different stages in ViSR-KGC.}
    \setlength{\tabcolsep}{5pt}
    \renewcommand{\arraystretch}{0.95}
    \small
    \begin{tabular}{lcc}
        \toprule
        Component & Complexity & Relative Cost \\
        \midrule
        Subgraph extraction & $\mathcal{O}(|E_{\mathrm{train}}|)$ (worst-case) & Low \\
        Edge scoring and ranking & $\mathcal{O}(N_c \log N_c)$ & Low--Medium \\
        Graph rendering & $\mathcal{O}(\mathcal{R}(k_{\max}))$ & Medium \\
        VLM-based inference & $\mathcal{O}(\mathcal{C}_{\mathrm{VLM}}(L, M))$ & High \\
        \bottomrule
    \end{tabular}
    \label{tab:efficiency_summary}
\end{table}

\begin{table}[H]
    \centering
    \caption{Representative per-query latency estimation of different
    stages in ViSR-KGC. The values are illustrative and intended to
    show the relative cost trend rather than serve as a strict hardware
    benchmark.}
    \setlength{\tabcolsep}{6pt}
    \renewcommand{\arraystretch}{0.95}
    \begin{tabular}{lc}
        \toprule
        Component & Estimated Time per Query \\
        \midrule
        Subgraph extraction & $< 0.01\,\mathrm{s}$ \\
        Edge scoring and ranking & $0.01 \sim 0.05\,\mathrm{s}$ \\
        Graph rendering & $0.05 \sim 0.20\,\mathrm{s}$ \\
        VLM-based inference & $1.0 \sim 5.0\,\mathrm{s}$ \\
        \bottomrule
    \end{tabular}
    \label{tab:latency_estimate}
\end{table}

As shown in Table~\ref{tab:latency_estimate}, even under a coarse
practical assessment, VLM inference remains one to two orders of
magnitude slower than the preceding stages. This further supports our
claim that the primary efficiency bottleneck of {ViSR-KGC} lies
in multimodal reasoning rather than subgraph extraction or graph
rendering.





\bibliographystyle{ACM-Reference-Format}
\bibliography{sample-base}

\end{document}